\documentclass[10pt,twocolumn,letterpaper]{article}
\usepackage{iccv}
\usepackage{times}
\usepackage{epsfig}
\usepackage{graphicx}
\usepackage{amsmath}
\usepackage{amssymb}
\usepackage{IEEEtrantools}
\usepackage{xfrac}
\usepackage{bm}
\usepackage{caption}
\usepackage{subcaption}
\usepackage{float}
\usepackage{color,soul}
\usepackage{xcolor}
\usepackage{mathtools}
\usepackage[titletoc,page]{appendix}

\usepackage{multirow}
\usepackage{authblk}
\usepackage[utf8]{inputenc}

\usepackage[pagebackref=true,breaklinks=true,letterpaper=true,colorlinks,bookmarks=false]{hyperref}

\renewcommand{\vec}[1]{\mathbf{#1}}
\newcommand{\mat}[1]{\mathbf{#1}}

\renewcommand{\eg}{e.g.\ }
\renewcommand{\wrt}{w.r.t.\ }
\newcommand*{\tran}{^{\mkern-1.5mu\mathsf{T}}}
\newcommand*{\negtran}{^{\mkern-1.5mu\mathsf{-T}}}

\definecolor{nicegreen}{HTML}{99C000}
\definecolor{nicegreen2}{HTML}{81d41a}
\definecolor{niceyellow}{HTML}{FDCA00}
\definecolor{niceorange}{HTML}{f5a300}
\definecolor{niceblue}{HTML}{0083cc}
\definecolor{nicepurple}{HTML}{a60084}

\DeclarePairedDelimiter\floor{\lfloor}{\rfloor}

\iccvfinalcopy 

\ificcvfinal\fi

\begin{document}
	
	\title{Temporally Consistent Horizon Lines}
	
	\author[1]{Florian Kluger}
	\author[1]{Hanno Ackermann}
	\author[2]{Michael Ying Yang}
	\author[1]{Bodo Rosenhahn}
	\affil[1]{Institut für Informationsverarbeitung, Leibniz Universität Hannover}
	\affil[ ]{{\tt\small \{kluger,ackermann,rosenhahn\}@tnt.uni-hannover.de}}
	\affil[2]{Scene Understanding Group, University of Twente}
	\affil[ ]{{\tt\small michael.yang@utwente.nl}}
	
	\maketitle
	\begin{abstract}
	    The horizon line is an important geometric feature for many image processing and scene understanding tasks in computer vision. 
	    For instance, in navigation of autonomous vehicles or driver assistance, it can be used to improve 3D reconstruction as well as for semantic interpretation of dynamic environments. 
	    While both algorithms and datasets exist for single images, the problem of horizon line estimation from video sequences has not gained attention. 
	    In this paper, we show how convolutional neural networks are able to utilise the temporal consistency imposed by video sequences in order to increase the accuracy and reduce the variance of horizon line estimates. 
	    A novel CNN architecture with an improved residual convolutional LSTM is presented for temporally consistent horizon line estimation.
	    We propose an adaptive loss function that ensures stable training as well as accurate results. 
	    Furthermore, we introduce an extension of the KITTI dataset which contains precise horizon line labels for 43699 images across 72 video sequences. 
	    A comprehensive evaluation shows that the proposed approach consistently achieves superior performance compared with existing methods.
	\end{abstract}
	\vspace{-.5em}
	\section{Introduction}
	Horizon lines are important low-level geometric image features that provide essential information about the relation between a 3D scene and the camera observing it. They can be used to infer the camera pose in form of a ground plane normal or a gravity vector, respectively.
	In autonomous driving, ground planes are often used to infer semantic properties of the dynamic environment~\cite{alvarez20103d,geiger2009monocular}. 
	Other applications include estimation of vanishing points~\cite{zhai2016detecting}, which provide information about the 3D structure of a scene, image metrology~\cite{criminisi2000single}, perspective correction~\cite{lee2012automatic} and camera pose estimation~\cite{ettinger2003vision, Hold-Geoffroy_2018_CVPR}. 
	\begin{figure}[t]
	\begin{center}
		\includegraphics[width=0.99\linewidth]{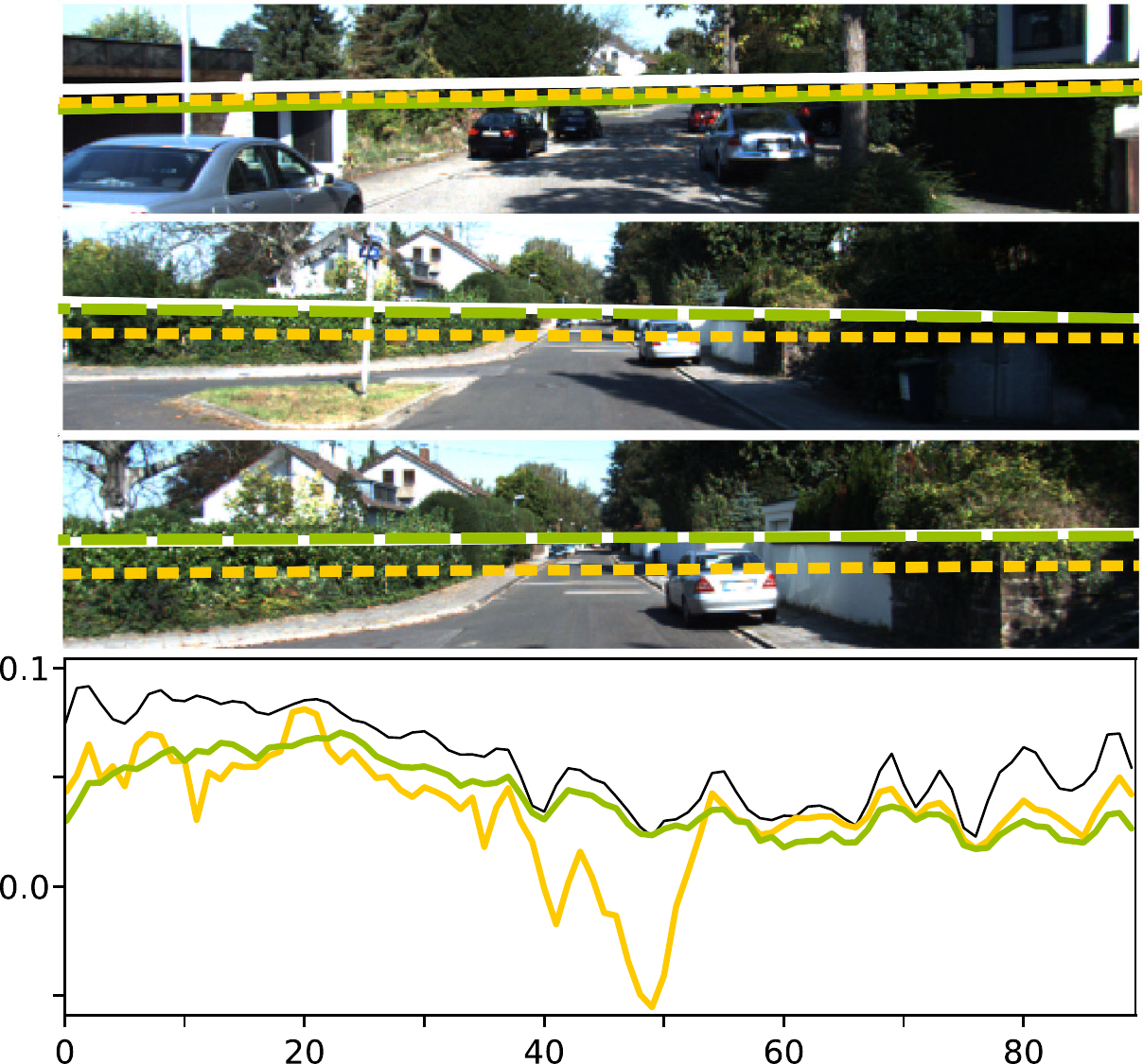}
	\end{center}
	\caption{Example sequence with our temporally consistent estimation in \textcolor{nicegreen}{green} (long dashes) and the best single frame algorithm in  \textcolor{niceyellow}{yellow} (short dashes). Ground truth in white/black. Top three rows: sample frames with horizon lines from the sequence. Bottom row: Horizon offset trajectory over time, best viewed in colour. The temporally consistent estimation is more accurate on average and contains fewer outliers.}
	\label{fig:teaser}
	\end{figure}
	
	For many applications, utilising temporal consistency has been demonstrated to improve performance. Examples include depth estimation~\cite{tananaev2018temporally}, motion segmentation~\cite{bertholet2018temporally}, action recognition~\cite{huang2018video}, super resolution~\cite{hanson2018bidirectional} and superpixel segmentation~\cite{reso2013temporally}. 
	Single image approaches for horizon line estimation may do gross mistakes when the image provides few or misleading clues. As illustrated by Fig.~\ref{fig:teaser}, an approach based on multiple images is less susceptible to these problems if it is able to transfer information from previous images of a sequence. 

	\subsection{Contributions}
	In this work, we present a novel approach for temporally consistent horizon line estimation based on a convolutional neural network combined with an improved convolutional long short-term memory (LSTM). A comprehensive evaluation demonstrates the ability of this approach to generate more accurate horizon line estimates with less variance.
	Since a na\"ive loss function does not track the geometric error of horizon lines very well, and a loss based on the geometric error exhibits singularities that may cause instability, we propose an adaptive loss function that combines both losses with a cosine annealing schedule. 
	This loss function yields significantly more accurate horizon estimates, yet ensures that the neural network training remains stable.
	In an ablation study, we investigate the influence of several hyper-parameters and architecture choices on the performance of the neural network models. 
	Furthermore, the \emph{KITTI Horizon} dataset is presented, an extension of the well established KITTI benchmark~\cite{Geiger2012CVPR}. 
	It contains accurate horizon line annotations for all video sequences of the KITTI dataset~\cite{Geiger2013IJRR}. 
	In summary, our \textbf{main contributions} are: 
	\begin{enumerate}
	    \item We present a novel CNN architecture for temporally consistent horizon line estimation based on an improved residual convolutional LSTM. 
	    \item We propose an adaptive loss function that yields accurate horizon line estimates and ensures stable training.
	    \item A large-scale video dataset for temporally consistent horizon line estimation, the \emph{KITTI Horizon} dataset.
	   	To the best of our knowledge, this is the first video dataset with accurate horizon line ground truth.
	\end{enumerate}

	\subsection{Types of Horizon Lines}
	\label{sec:horizon_types}
	It is possible to distinguish three types of horizon lines: the \emph{visible horizon}, the \emph{true horizon} and the \emph{geometrical horizon}. The visible horizon is the apparent line which separates earth and sky. Its appearance is often shaped by the surroundings of an observer in the presence of entities like mountains, buildings or trees. If the view of an observer is unobstructed -- at sea, for example -- the visible horizon becomes identical to the true horizon. Assuming a spherical earth surface, the true horizon is the projection of a circle containing all points on the earth which are tangent to light rays passing through the point of view of an observer.
	
\begin{figure}[t]
	\begin{center}
		\includegraphics[width=0.8\linewidth]{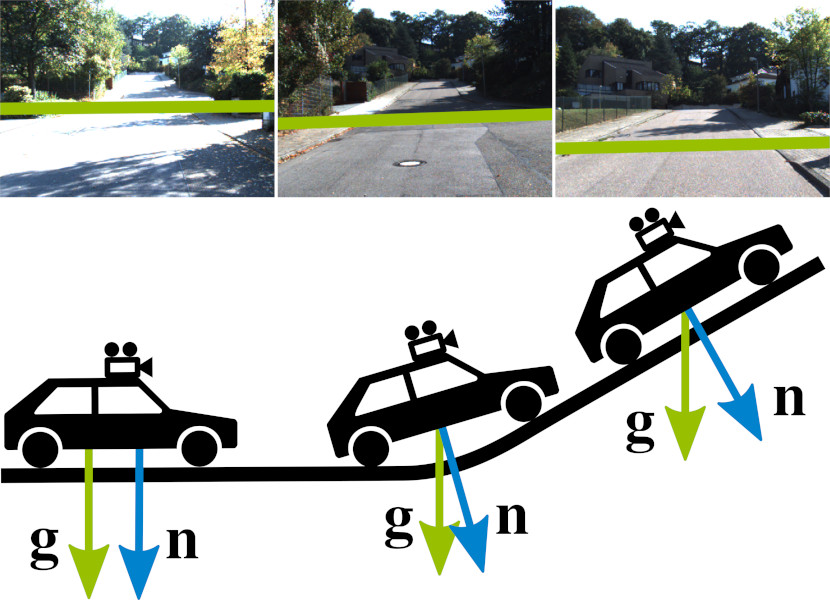}
	\end{center}
	\caption{Cropped images from a KITTI sequence with annotated horizon lines (top), and a sketch of the trajectory of the car with gravity vector $\vec{g}$ and plane normal $\vec{n}$ (bottom).
	}
	\label{fig:gravity_and_planes}
	\end{figure}

\begin{figure}[b]
	\begin{center}
		\includegraphics[width=0.7\linewidth]{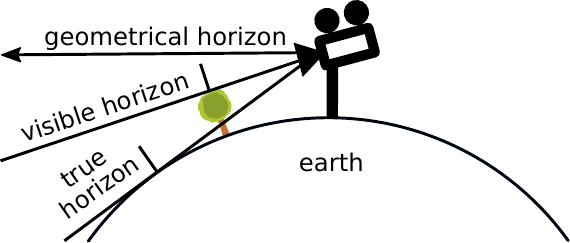}
	\end{center}
	\caption{Illustration of horizon line types (Sec.~\ref{sec:horizon_types}).}
	\label{fig:earth_and_horizons}
	\end{figure}

	The geometrical horizon $\vec{h}$ is defined as the vanishing line, i.e. the projection of the line at infinity, for any plane orthogonal to the local gravity vector $\vec{g}$:
	\begin{equation}
	\vec{h} \propto \mat{K}\negtran \mat{R} \vec{g} \quad ,
	\label{eq:gravity_to_horizon}
	\end{equation}
	with $\mat{R}$ being the orientation and $\mat{K}$ being the intrinsic calibration of the camera. 
	Without loss of generality, we assume that $\vec{g} \propto \begin{bmatrix} 0, & 1, & 0 \end{bmatrix}\tran $ is parallel to the the zenith direction.
	As illustrated by Fig.~\ref{fig:gravity_and_planes}, the geometrical horizon is generally not identical to the vanishing line of the plane an observer is standing on, as its normal vector may not be parallel to $\vec{g}$, \eg when located on an incline. Being a theoretical construction, the geometrical horizon is imperceptible to an observer. 
	However, given the intrinsic calibration $\mat{K}$, knowledge of the geometrical horizon is sufficient to estimate camera tilt and roll \wrt a global coordinate system. 
	Fig.~\ref{fig:earth_and_horizons} illustrates the conceptual differences between the three horizons.	
	Since the remainder of this paper considers the geometrical horizon, it will be simply referred to as the horizon from hereon. 
	
	
	\subsection{Related Work}
	\label{sec:related_work}
	In the past, numerous approaches for horizon line estimation have been proposed, and they can be differentiated into a number of categories. 
	Most methods rely on vanishing points (VPs)~\cite{barinova2010geometric,kluger2017deep,kovsecka2002video,lezama2014finding,Simon_2018_ECCV,tardif2009non,vedaldi2012self,wildenauer2012robust,xu2013minimum} which they detect by grouping oriented elements like line segments or edges into clusters which have the same orientation in 3D space. 
	If at least two vanishing points are known, the horizon line can be derived. 
	Some of these methods~\cite{kovsecka2002video,rother2002new,tardif2009non,wildenauer2012robust} rely on the Manhattan-world assumption~\cite{coughlan1999manhattan}, \ie they restrict their solution space to three VPs of orthogonal directions and are hence applicable to only a limited number of scenes. 
	Others~\cite{barinova2010geometric,lezama2014finding,schindler2004atlanta,Simon_2018_ECCV,xu2013minimum} use the more permissible Atlanta-world assumption~\cite{schindler2004atlanta}, which expects all horizontal VPs to be of orthogonal direction to a zenith VP. 
	This assumption is still restrictive, as it does not cover scenes which contain planes that are oblique to a defined zenith. 
    Several of the aforementioned methods consider two benchmark datasets in their evaluation: the York Urban Dataset~\cite{denis2008efficient} (YUD) and the Eurasian Cities Dataset~\cite{barinova2010geometric}. 
    Both are relatively small  and of limited diversity \wrt the types of scenes they depict. 
    In 2016, Workman et al.~\cite{workman2016hlw} presented the Horizon Lines in the Wild (HLW) dataset, which contains horizon line ground truth for 100553 images taken at various locations. 
    Availability of such a large-scale dataset has lead to an emergence of deep-learning based algorithms~\cite{Lee_2017_ICCV,workman2016hlw,zhai2016detecting} more recently. 
    Workman \etal~\cite{workman2016hlw} present a convolutional neural network (CNN) which directly estimates the horizon line from a single image, formulated as either a regression or a classification task. 
    Lee \etal~\cite{Lee_2017_ICCV} use a different approach: they randomly sample lines within the image borders and feed them, along with the image, into a CNN which incorporates their proposed \emph{line pooling layer}. 
    This CNN then provides a classification whether the sampled line is the horizon of the image and, in addition, computes refined line coordinates. 
    The method of Zhai \etal~\cite{zhai2016detecting} is a hybrid approach. 
    It uses a CNN, similar to~\cite{workman2016hlw}, to predict a horizon line, but then jointly optimises its location together with VPs which are estimated based on line segments that have been detected in a preprocessing step.
	All these works have in common that they target the problem of \emph{single image} horizon line estimation. 
	To the best of our knowledge, general datasets and algorithms targeted specifically at horizon line estimation from \emph{video sequences} do not exist.

	\section{KITTI Horizon Dataset}
	\label{sec:kitti_horizon_dataset}
	We introduce the \emph{KITTI Horizon Dataset}, a new addition to the KITTI raw dataset~\cite{Geiger2013IJRR} with accurate horizon line annotations for all video sequences. 
	
	\subsection{Limitations of Existing Datasets}
	Three datasets have been commonly used for horizon line estimation in recent years: the York Urban Dataset~\cite{denis2008efficient} (YUD), the Eurasian Cities Dataset~\cite{barinova2010geometric} (ECD) and Horizon Lines in the Wild~\cite{workman2016hlw} (HLW). 
	YUD is a relatively small dataset of 102 images depicting in- and outdoor scenes within a confined area, taken with the same camera under similar conditions. 
	While ECD is somewhat more diverse than YUD, it is still very small with just 103 images. 
	HLW, on the other hand, is significantly larger and contains 100553 images, making it much better suited for data-intensive deep learning approaches. 
	Unlike YUD and ECD, HLW was not labelled manually, but in an automatic process using structure from motion. 
	It appears, however, that this process has limited precision, as some images in HLW have clearly inaccurate horizon line labels. 
	Beyond that, all three datasets have in common that they do not contain video sequences, which means that they can only be used for \emph{single image} horizon line estimation and are ill-suited for research on \emph{temporally consistent} horizon line estimation. 
	To our knowledge, the Singapore Maritime Dataset~\cite{prasad2017video} (SMD) is the only video dataset with annotated horizon lines. 
	Although it is relatively large, containing 21981 annotated frames, its diversity is very limited since it exclusively shows maritime scenes of similar appearance. 
	More importantly, however, the horizon labels in SMD describe the \emph{true horizon} as opposed to the \emph{geometrical horizon}. 
	Consequently, a new dataset is needed for temporally consistent geometrical horizon line estimation.
	\subsection{KITTI}
	KITTI~\cite{Geiger2013IJRR} is a computer vision dataset which was captured using a sensor array mounted on top of a vehicle.
	Sensors used for the recordings include four front-facing video cameras and a high accuracy inertial measurement unit (IMU), among others. 
	Several benchmarks for various applications, such as object detection, depth estimation or semantic segmentation, have been published~\cite{ Geiger2012CVPR}. 
	For horizon line estimation such a benchmark does not exist. We can, however, compute accurate horizon line ground truth using the IMU data provided by KITTI, at no additional cost.

	\begin{figure*}
		\begin{center}
			\includegraphics[width=0.33\linewidth]{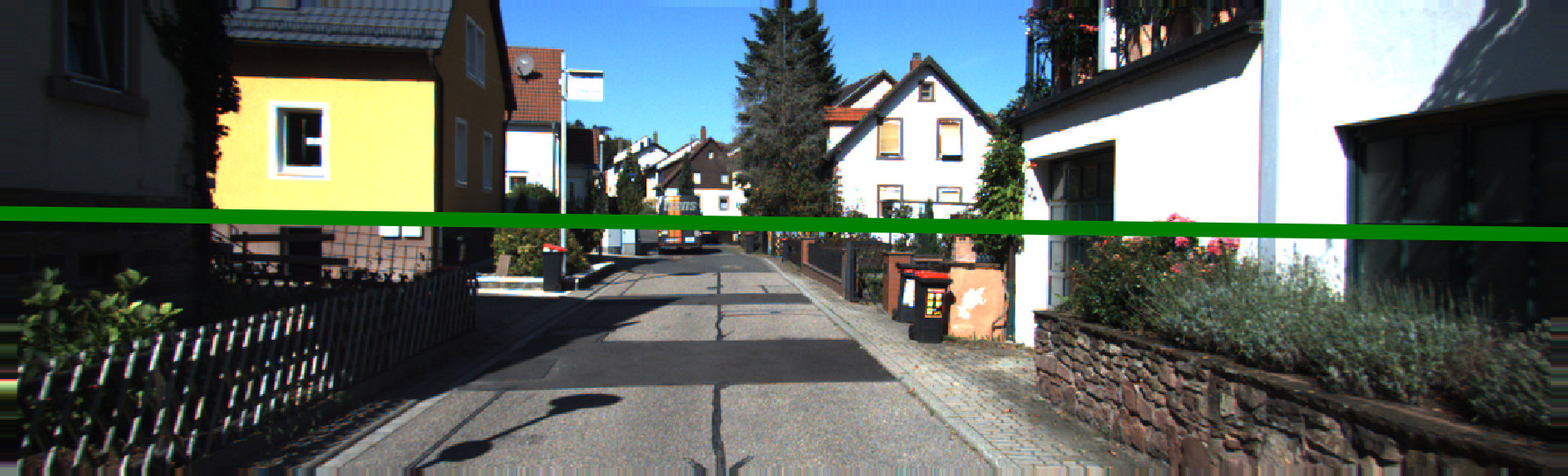}
			\includegraphics[width=0.33\linewidth]{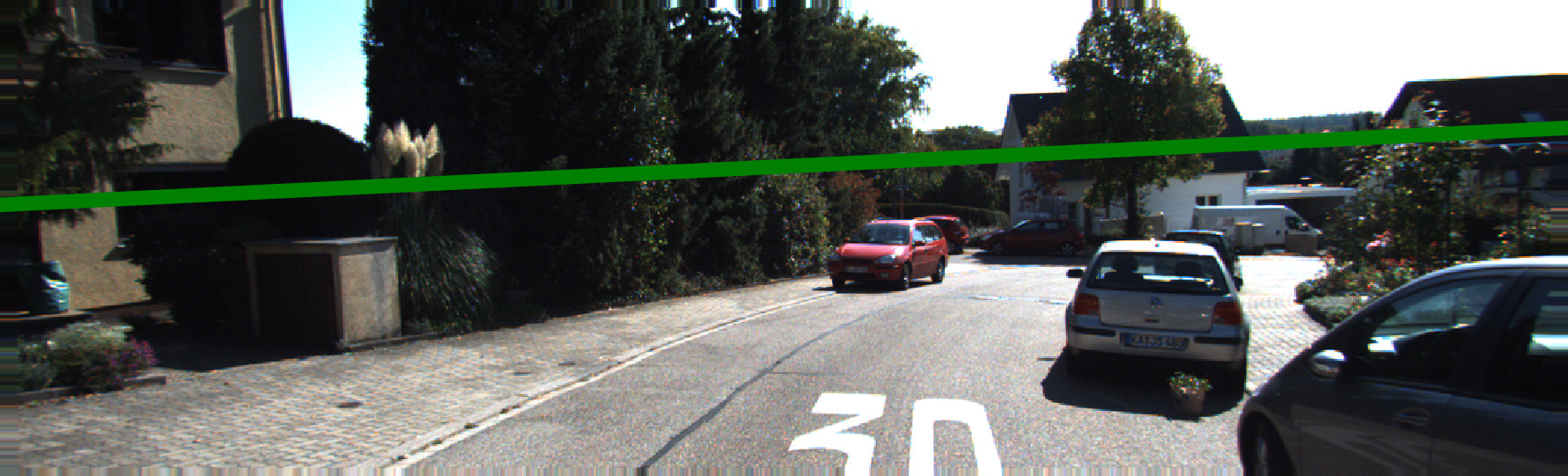}
			\includegraphics[width=0.33\linewidth]{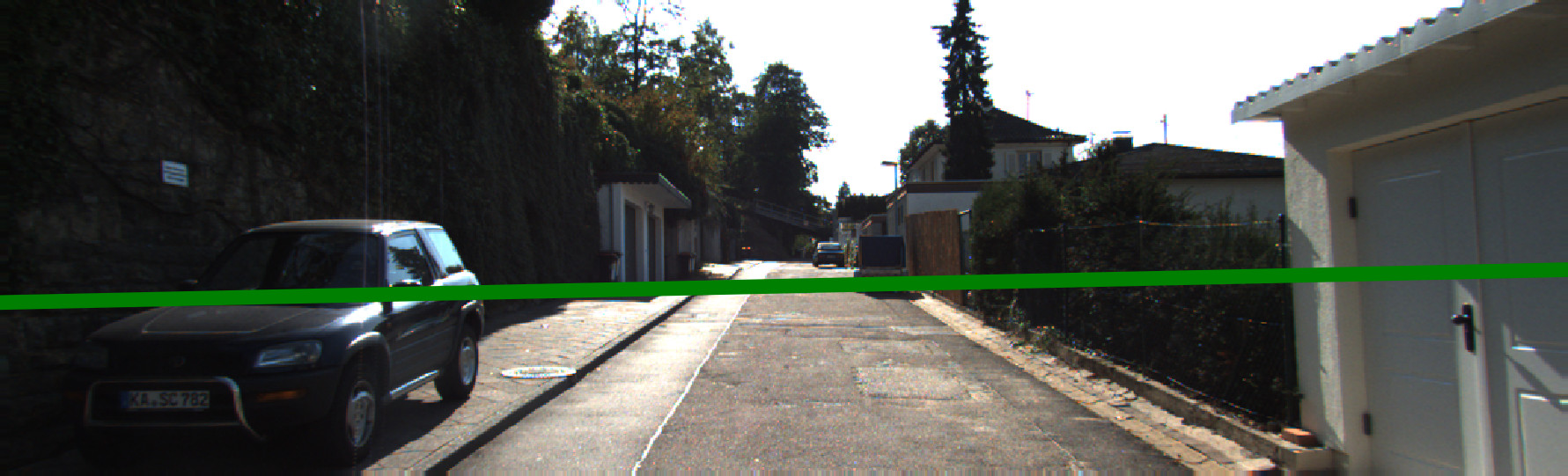}
		\end{center}
		\caption{Example frames from KITTI with annotated horizon line.}
		\label{fig:kitti_examples}
	\end{figure*}
	
	\subsection{Horizon Line Ground Truth}
	KITTI provides an accurate absolute pose $\mat{R}_{\mathsf{IMU}}$ of the IMU in 3D space for every image. 
	Together with the relative pose between IMU and camera $N \in \{1,2,3,4\}$, $\mat{R}_{\mathsf{IMU}\rightarrow N}$, we can compute the normalised gravity vector $\vec{g}_N \propto \mat{R}_{\mathsf{IMU}\rightarrow N} \mat{R}_{\mathsf{IMU}} \begin{bmatrix}0, & 1, & 0\end{bmatrix}\tran $  in the coordinate system of the camera.

	As explained in Sec.~\ref{sec:horizon_types}, the projection of a gravity vector $\vec{g}$ into the camera using Eq.~\ref{eq:gravity_to_horizon} yields the horizon line in homogeneous coordinates:
	\begin{equation}
		\vec{h}_N \propto \mat{K}_N^{-\mathsf{T}}  \mat{R}_{\mathsf{IMU}\rightarrow N} \mat{R}_{\mathsf{IMU}} 
		\begin{bmatrix}
		0, & 1, & 0
		\end{bmatrix}\tran
		\, .
	\end{equation}
	As this process requires no manual labelling or other human intervention, we can compute the ground truth horizon for all images fully automatically. 
	Fig.~\ref{fig:kitti_examples} shows a few examples. 
	In the left-hand image, the ground plane appears nearly perpendicular to the gravity vector, hence the horizon line is virtually identical to the vanishing line of that ground plane. 
	In the other two images, however, they are clearly distinct due to the fact that the ground plane is sloping downwards (middle image) or upwards (right-hand image).
	
	\subsection{Train, validation and test split}
	The complete published KITTI dataset consists of 47962 frames across 157 sequences. Several sequences show the same stationary scene, and only differ \wrt the people walking across the image. 
	As these are of negligible value for our task, we discarded all but one, so that 72 sequences with 43699 frames remain. 
	As no official split exists for the raw dataset, we divided the video sequences into roughly 70\% training, and 15\% validation and test data each. 
	Care was taken to ensure that sequences showing very similar scenes, \eg the same intersection, do not end up in different parts of the split. 
	As there is a strong imbalance in sequence length, \eg some sequences contain less than 100 frames and others have several thousand, we divided one of the longer videos equally and put it into the test and validation sets.
	
	\section{Single Image Estimation}
	\label{sec:single_image_estimation}
	We obtained the source code of recent single image algorithms~\cite{kluger2017deep,lezama2014finding,Simon_2018_ECCV,workman2016hlw,zhai2016detecting}. 
	In addition, we compare our own single image algorithm along with these methods. Thereby, we obtain a detailed and unbiased comparison that clearly highlights the features of our temporally consistent approach.
	Our single image algorithm is based on a CNN, similar to the regression approach presented in~\cite{workman2016hlw}. 
	We parametrise the horizon line $\vec{h}$ by offset $\omega$ and slope $\theta$. With $W$ being the image width, its representation in homogeneous coordinates is defined as:
	\begin{equation}
	\label{eq:horizon_definition}
	    \vec{h}(\omega,\theta) = 
	    \begin{bmatrix}
	    \sin{\theta}, &   \cos{\theta}, &  -\frac{W}{2}\sin\theta - \omega \cos\theta
	    \end{bmatrix}\tran
	    \, .
	\end{equation}
	We replace the GoogleNet~\cite{szegedy2015going} of~\cite{workman2016hlw} with the more recent and efficient ResNet~\cite{he2016deep}, and use the most shallow 18-layer variant (ResNet18). 
	The classification layer of the ResNet is replaced by two fully connected layers with single real valued outputs for $\omega$ and $\theta$. 
	Apart from downscaling the image, we do not perform any pre- or post-processing.
	
	\begin{figure*}[]
	\begin{center}
		\includegraphics[width=0.99\linewidth]{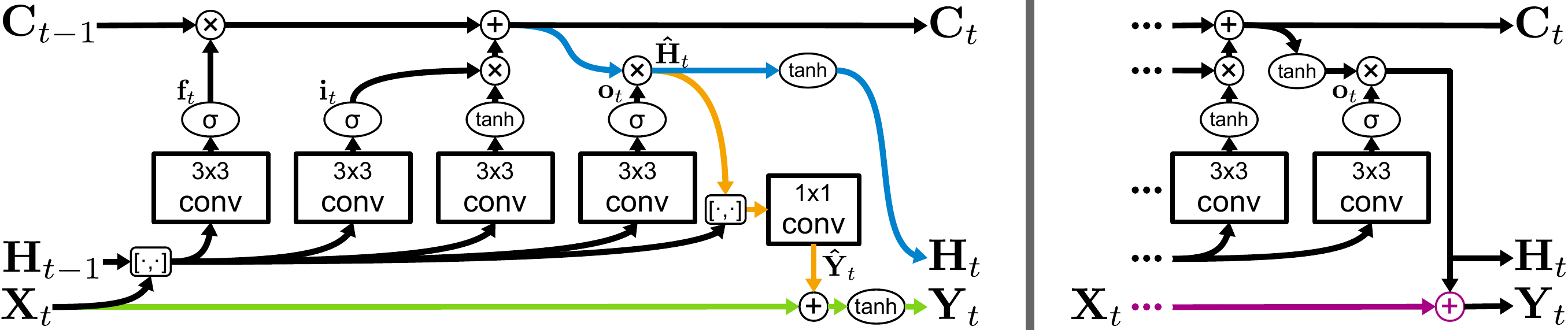}
	\end{center}
	\caption{
	ConvLSTM with residual paths as described in Sec.~\ref{sec:residual_convlstm}. The $[\cdot,\cdot]$-operator denotes concatenation along the channel axis.  \emph{Left}: Our proposed ConvLSTM with residual paths and dense connections.  Changes \wrt~a standard ConvLSTM: residual connection from $\mat{X}_t$ to $\mat{Y}_t$ in \textcolor{nicegreen2}{green}; dense connection from $\mat{X}_t$, $\mat{\hat{H}}_t$ and $\mat{H}_{t-1}$ to $\mat{Y}_t$ in \textcolor{niceorange}{orange}; reversal of operation order in \textcolor{niceblue}{blue}. \emph{Right}: Part of a standard ConvLSTM, with a na\"ive implementation of a residual connection in \textcolor{nicepurple}{purple}.}
	\label{fig:convlstm}
    \end{figure*}
    
	\section{Temporally Consistent Estimation}
	\label{sec:temporally_consistent_estimation}
	
	Possibly the simplest way to utilise the temporal consistency of video sequences is applying a single-frame algorithm first, and then averaging the results. 
	For online applications, a reasonable choice of filter would be the exponential moving average, or exponential smoothing filter~\cite{brown2004smoothing}. Given a sequence $x_t$, the output of the filter is defined as:
	\begin{equation}
	\label{eq:exp_smooth}
	    s_t = \alpha x_t + (1-\alpha)s_{t-1}
	\end{equation}
	While easy to implement, it always just achieves a compromise between suppressing noise and outliers, and preserving actual trajectory changes. 
	Bai \etal~\cite{bai2018empirical} propose temporal convolutional networks (TCN), an extension of regular CNNs by causal convolutions~\cite{oord2016wavenet} along an additional temporal dimension. 
	Across time, the TCN has a fixed field of view which limits the sequence length along which it is able to infer correlations. 
	We therefore chose to investigate an approach based on long-short term memory (LSTM)~\cite{hochreiter1997long}. 
	We devised a novel approach combining the ResNet~\cite{he2016deep} architecture with an improved convolutional LSTM layer. 
	
	\subsection{Convolutional LSTM}
    LSTM cells are a particular type of recurrent neural network (RNN) that have been proven effective in modelling both long- and short-term dependencies of sequential data~\cite{graves2013hybrid,sak2014long,wu2016google}.
	The convolutional LSTM (ConvLSTM)~\cite{tananaev2018temporally,xingjian2015convolutional} is a variant that operates on 3D tensors instead of vectors and replaces all matrix multiplications with kernel convolutions. 
	Given a sequence of inputs $\mat{X}_1, \dots, \mat{X}_t$, the cell state $\mat{C}_t$ and hidden state $\mat{H}_t$ of a ConvLSTM can be computed as follows, where '$*$' is the convolution operator and '$\odot$' denotes the Hadamard product: 
	\begin{IEEEeqnarray}{rCl}
	\IEEEyesnumber
	\label{eq:convlstm_i}
	\mat{i}_t &=& \sigma(\mat{W}_{xi} * \mat{X}_t + \mat{W}_{hi} * \mat{H}_{t-1} + \mat{b}_i ) \\
	\label{eq:convlstm_f} \mat{f}_t &=& \sigma(\mat{W}_{xf} * \mat{X}_t + \mat{W}_{hf} * \mat{H}_{t-1} + \mat{b}_f ) \\
	\label{eq:convlstm_o}
	 \mat{o}_t &=& \sigma(\mat{W}_{xo} * \mat{X}_t + \mat{W}_{ho} * \mat{H}_{t-1} + \mat{b}_o ) \\
	\mat{C}_t &=& \mat{f}_t \odot \mat{C}_{t-1} \, + \nonumber\\
	\label{eq:convlstm_C} 
	          && \mat{i}_t \odot \tanh( \mat{W}_{xc} * \mat{X}_t + \mat{W}_{hc} * \mat{H}_{t-1} + \mat{b}_c )\\
    \label{eq:convlstm_H}
	 \mat{H}_t &=& \mat{o}_t \odot \tanh(\mat{C}_t)
    \end{IEEEeqnarray}

	The hidden state is usually treated as the output of the cell, \ie $\mat{Y}_t = \mat{H}_t$. Variants with additional connections~\cite{gers2000recurrent}, or other activation functions~\cite{tananaev2018temporally} exist as well.
    
	\subsection{Residual Convolutional LSTM} 
	\label{sec:residual_convlstm}
	We propose an improved convolutional LSTM structure that incorporates both residual and dense connections. 
	As previous works~\cite{he2016deep, wu2016google} have shown, residual connections improve gradient flow in deep neural networks, which makes them easier and faster to train. 
	He \etal~\cite{he2016deep} integrated residual connections into a CNN. If we consider a shallow stack $l$ of convolutional layers performing an operation $F_l(\mat{x})$ on an input $\mat{x}_{l-1}$, the output $\mat{x}_l$ of such a stack is:
	 $   \mat{x}_l = g(F_l(\mat{x}_{l-1}) + \mat{x}_{l-1}) \, $,
	with $g(\cdot)$ being a nonlinear activation function, \eg $\mathrm{ReLU}$. 
	In~\cite{wu2016google}, this idea was applied to a network of stacked LSTM cells. 
	Each LSTM cell computes a hidden state $\vec{h}_t$ and a cell state $\vec{c}_t$ based on an input $\mat{x}_t$ and the states at the previous time step:
	$    \vec{h}_t, \vec{c}_t = \mathrm{LSTM}(\vec{h}_{t-1}, \vec{c}_{t-1}, \vec{x}_t) \, $.
	A residual connection is then applied to generate the final output of the layer:
	\begin{equation}
	\label{eq:naive_residual}
	    \vec{y}_t = \vec{h}_t + \vec{x}_t \, .
	\end{equation}
	In this case, the non-linearity $g(\cdot)$ is part of the LSTM, \ie it is applied before the residual connection. The notion of improving information flow through a neural network via connections that skip a number of layers was implemented in yet another manner by Huang \etal~\cite{huang2017densely}. In their DenseNet CNN architecture, feature-maps of $M$ preceding layers $\mat{x}_{l-M}, \dots \mat{x}_{l-1}$ are concatenated channel-wise and fed into the current layer $F_l(\mat{x})$:
	$    \mat{x}_l = g(F_l([\mat{x}_{l-M}, \dots, \mat{x}_{l-1}]) \, $.
	In order to arrive at our improved ConvLSTM, we combine the aforementioned principles and incorporate them as follows. Fig.~\ref{fig:convlstm} illustrates our proposed structure on the left side, while the right side shows the standard ConvLSTM with a na\"ive residual connection as per Eq.~\ref{eq:naive_residual} for comparison. In keeping with the original ResNet definition, we define a residual connection between input and output:
	\begin{equation}
	    \mat{{Y}}_t = \tanh(\mat{\hat{Y}}_{t} + \mat{X}_{t}) \, .
	\end{equation}
	As Eq.~\ref{eq:convlstm_H} shows, the hidden state $\vec{H}_t$ amounts to a masked cell state $\vec{C}_t$. 
	We argue that this inhibits the flow of information from both $\vec{X}_t$ and $\vec{H}_{t-1}$ to the output $\vec{Y}_t$. 
	Normally, information must pass through Eqs.~\ref{eq:convlstm_i}-\ref{eq:convlstm_H} and thus through $\vec{C}_t$ before it eventually reaches $\vec{Y}_t$. 
	We therefore introduce an additional convolutional layer into the ConvLSTM, which directly takes the concatenation of $\vec{X}_t$, $\vec{H}_{t-1}$ and an intermediate hidden state $\vec{\hat{H}}_{t}$ as an input, similar to the way convolution layers in DenseNet operate, in order to produce an intermediate output $\vec{\hat{Y}}_t$:
		\begin{equation}
	    \mat{\hat{Y}}_t = \mat{W}_{xy} * \mat{X}_t + \mat{W}_{hy} * \mat{H}_{t-1} + \mat{W}_{\hat{h}y} * \mat{\hat{H}}_t \, .
	\end{equation}
	Finally, in order to avoid application of the $\tanh$ activation twice onto the information from $\mat{C}_t$, we switch the order of operation in Eq.~\ref{eq:convlstm_H}, \ie:
	\begin{IEEEeqnarray}{rCl}
	\IEEEyesnumber
	\label{eq:convlstm_skip_Hhat}
	    \mat{\hat{H}}_t &=& \mat{o}_t \odot \mat{C}_t \, , \\
	\label{eq:convlstm_skip_H}
	    \mat{{H}}_t &=& \tanh(\mat{\hat{H}}_t) \, .
	\end{IEEEeqnarray}
	
	\subsection{Horizon Line Estimation Network}
	\label{sec:horizon_line_network}
	\begin{figure*}
	\begin{center}
		\includegraphics[width=0.99\linewidth]{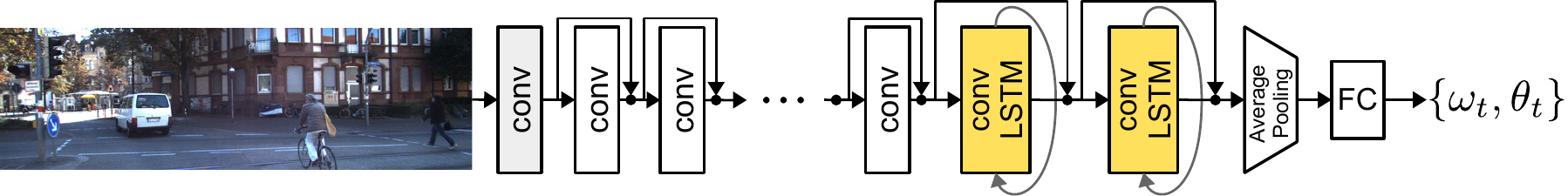}
	\end{center}
	\caption{Proposed neural network structure employing ConvLSTM layers as described in Sec.~\ref{sec:horizon_line_network}. 
	Two ConvLSTM layers are inserted between the last convolutional layer and the global average pooling layer of our ResNet18-based CNN.
	The outputs $\omega$ and $\theta$ are the offset and slope, respectively.
	}
	\label{fig:convlstmnet}
    \end{figure*}
    We expand our single image CNN described in Sec.~\ref{sec:single_image_estimation} with our modified ConvLSTM presented in Sec.~\ref{sec:residual_convlstm} in order to create a temporally consistent architecture. 
    As Fig.~\ref{fig:convlstmnet} shows, two ConvLSTM layers are inserted between the last convolutional layer and the global average pooling layer of our ResNet18-based CNN. 
    Intuitively, applying the ConvLSTM at this stage makes most sense, as we would expect it to find temporal correlations between higher-level features which are most pertinent to the task of horizon estimation.
    
	\subsection{Loss Function}
	\label{sec:loss_function}
	Our CNN has two real valued outputs: offset $\omega$ and slope $\theta$ of the predicted horizon line. 
	We compute two loss terms; the first one is the Huber loss of $\omega$ and $\theta$ computed \wrt the ground truth; the second one is the maximum horizon error within the image. 
	Combining these two losses allows us to benefit from a gain in accuracy elicited by minimising the maximum horizon error, while avoiding the instability it can cause.
	The Huber loss~\cite{huber1992robust} is defined as:
	$$    L_H(x,\hat{x}) =   
	    \begin{cases} 
	        \frac{1}{2}(x-\hat{x})^2             & \text{for } \left|x-\hat{x}\right| \leq 1 \, , \\ 
	        \left|x-\hat{x}\right| - \frac{1}{2} & \text{otherwise.}
        \end{cases}$$
	We define the first loss term as the Huber loss of $\omega$ and $\theta$ computed \wrt the ground truth $\hat{\omega}$ and $\hat{\theta}$:
	\begin{equation}
	\label{eq:omega_theta_loss}
	    L_{\omega,\theta} = L_H(\omega, \hat{\omega}) + L_H(\theta, \hat{\theta}) \, .
	\end{equation}
	As this loss term does not exactly track the maximum horizon error, which is the quantity we actually seek to minimise, we have defined a second loss term.
	The maximum horizon error is defined as the maximum distance between the estimated horizon $\vec{h}(\omega,\theta)$, as defined by Eq.~\ref{eq:horizon_definition}, and the ground truth horizon $\vec{h}(\hat{\omega},\hat{\theta})$ between the left- and rightmost borders of the image, normalised to image height $H$. 
	The $y$-coordinate of the intersection of $\vec{h}$ with a vertical line at $x$ is determined by:
    \begin{equation}
    \label{eq:y_omega_theta_x}
        y(\omega, \theta, x) = \left(x-\frac{W}{2}\right) \tan\theta - \omega \, .
    \end{equation}
	Let $d_{y,0}$ and $d_{y,W}$ be the left- and right-most distances between the two horizons, 
	$    d_{y,x} = \left| y(\omega,\theta,x) - y(\hat{\omega},\hat{\theta},x)\right| \, $.
	The maximum horizon error  $L_e$ can then be defined as:
	\begin{equation}
	    \label{eq:max_horizon_error}
	    L_e =   
	    \begin{cases} 
	        \frac{1}{H}d_{y,0} & \text{for } d_{y,0} \geq d_{y,W} \, , \\ 
	        \frac{1}{H}d_{y,W} & \text{otherwise.}
        \end{cases}
	\end{equation}
	While $L_e$ directly reflects the quantity we aim to minimise, it contains singularities for $\theta = (\sfrac{\pi}{2} + n\pi), n \in \mathbb{N}$, due to the $\tan \theta$ term in Eq.~\ref{eq:y_omega_theta_x}. 
	This causes $L_e$ to become excessively large if $\theta$ is poorly estimated, which may be the case especially at the beginning of neural network training. 
	We therefore use only $L_{\omega,\theta}$ at first, when estimates are still very inaccurate and noisy, and gradually switch over to  $L_e$ on a cosine schedule similar to~\cite{loshchilov2016sgdr}. 
	With $t$ being the current epoch and $T$ being the maximum number of epochs, the schedule is defined by:
	 $   \lambda(t) = \frac{1}{2} + \frac{1}{2}\cos\left(\pi \cdot \frac{t}{T}\right) \, .$
	Using this, the final loss $L$ is defined as:
	\begin{equation}
	    L(t) = \lambda(t) \cdot L_{\omega,\theta} + (1-\lambda(t)) \cdot L_e \, . 
	\end{equation}
	
	\section{Experiments}

	\begin{table*}
		\begin{center}
			\begin{tabular}{|ll|cc|cc|cc|cc|}
				\cline{3-10}
				\multicolumn{2}{c}{}&
				\multicolumn{2}{|c}{AUC (horizon)} & 
				\multicolumn{2}{|c|}{MSE $\times 10^{-3}$ } & 
				\multicolumn{2}{|c|}{$A_{TV} \times 10^{-3}$ } & 
				\multicolumn{2}{|c|}{AUC (pose) }\\
				\cline{3-10}
				\multicolumn{2}{c|}{} & 
				val & test &  
				val & test &  
				val & test & 
				val & test \\
				\hline
				\multicolumn{2}{|l|}{Lezama \etal~\cite{lezama2014finding}}&    34.17\%& 30.45\%& 
				                                                                \textgreater1000 & \textgreater1000 &
				                                                                2397  & 1537  & 
				                                                                28.79\%& 25.28\%\\
				\hline
				\multicolumn{2}{|l|}{Kluger \etal~\cite{kluger2017deep}}   &    54.27\%& 48.21\%& 
				                                                                \textgreater1000 & \textgreater1000 & 
				                                                                188.6& 206.4 &
				                                                                47.29\%&41.47\%  \\				
				\hline
				\multicolumn{2}{|l|}{Simon \etal~\cite{Simon_2018_ECCV}}&   57.03\%& 47.84\%& 
				                                                            84.26 & 224.0 &
				                                                            65.94 & 88.71 & 
				                                                            50.97\%& 41.98\%\\
				\hline
				\multicolumn{2}{|l|}{Zhai \etal~\cite{zhai2016detecting}}&  60.97\%& 50.98\%& 
				                                                            \textgreater1000 & \textgreater1000 &
				                                                            91.56 & 1575  & 
				                                                            53.52\%& 43.47\%\\
				\hline
				\multicolumn{2}{|l|}{Workman \etal~\cite{workman2016hlw}}&  70.32\%& 66.48\%&
				                                                            9.208& 11.19&
				                                                            6.893 & 8.430 & 
				                                                            62.36\%& 58.58\%\\
				\hline
				\hline
				\multicolumn{2}{|l|}{\emph{Average} baseline}&  69.40\%& 64.18\%&
				                                                8.800 &  12.20&
				                                                6.091 & 5.123 & 
				                                                59.45\%& 54.98\%\\
				\hline
				\hline
				 \textbf{single frame}& trained on HLW& 
				                                            71.10\% & 63.64\%& 
				                                            10.41 & 14.31 &
				                                            13.90 & 15.71 & 
				                                            64.02\%& 55.20\%\\
				                                            \cline{2-10}
				  (Sec.~\ref{sec:single_image_estimation})&trained on KITTI-H&  
				                                    
				                                    77.42\% & 74.08\%&
				                                               6.024 & 7.025&
				                                    5.061 & 5.585 & 
				                                    70.51\%& 66.62\%\\
				&  w/ exp. smoothing    & 77.44\% & 74.11\% 
				                        & 5.986 & 6.987 
				                        & 4.337 & 4.687
				                        & 70.50\%& 66.64\%\\
				\hline
				TCN~\cite{bai2018empirical} (3-3-5) & 
				                        & 75.42\%& 71.80\%
				                        & 6.392& 8.318
				                        & 4.945& \textbf{4.937}
				                        & 67.33\%& 64.21\%\\
				\hline
				\multicolumn{2}{|l|}{\textbf{temporally consistent} (Sec.~\ref{sec:temporally_consistent_estimation})}
   												                        & \textbf{78.09\%}& \textbf{74.55\%}& 
				                                                        \textbf{5.427} & \textbf{6.731} &
												                        \textbf{4.619} & {4.984} & 
												                        \textbf{71.17\%}& \textbf{67.33\%}\\
				& w/ exp. smoothing &   \textbf{78.11\%}& \textbf{74.68\%}& 
				                        \textbf{5.405} & \textbf{6.712} &
				                        \textbf{4.159} & \textbf{4.404} & 
										\textbf{71.19\%}& \textbf{67.49\%}\\

				\hline
			\end{tabular}
		\end{center}
		\caption{Horizon estimation results on the KITTI Horizon (Sec.~\ref{sec:kitti_horizon_dataset}) validation and test sets using the metrics described in Sec.~\ref{sec:evaluation_metrics}. AUC: higher is better; MSE and $A_{TV}$: lower is better. Refer to Sec.~\ref{sec:kitti_horizon_results} for a detailed discussion.}
		\label{tab:kitti_results}
	\end{table*}	
    
	We empirically demonstrate the effectiveness of our temporally consistent horizon line estimation pipeline on the KITTI Horizon validation and test sets and compare it with state-of-the-art single-image algorithms and other temporally consistent baselines. Additional ablation studies show the importance of individual parts of this pipeline for achieving these results.
	
	\subsection{Implementation Details}
	We implemented the proposed neural network architectures using PyTorch~\cite{paszke2017automatic}. 
	On KITTI Horizon, all networks were trained for 160 epochs with stochastic gradient descent using a cosine annealing learning rate schedule~\cite{loshchilov2016sgdr} starting at $10^{-1}$ and ending at $10^{-3}$. 
	Training was repeated four times with different random seeds, and the model with the highest validation AUC chosen. 
	We downscale each image by a factor of two and apply cutout~\cite{devries2017cutout}, colour jitter, random rotations and random shifts for data augmentation. 
	We initialise the weights of the first nine convolutional layers of the networks from a ResNet18 pretrained on ImageNet~\cite{imagenet_cvpr09} while other layers are initialised randomly. 
	Training batches always contain $B$ sequences of $S$ consecutive frames from the KITTI Horizon training set, and batch size $B$ and sequence length $S$ were set to fulfil $S\cdot B = 128$.
	
    \subsection{Evaluation Metrics}
    \label{sec:evaluation_metrics}
    As in~\cite{kluger2017deep,lezama2014finding,Simon_2018_ECCV,workman2016hlw,zhai2016detecting}, we compute the maximum horizon error defined in Eq.~\ref{eq:max_horizon_error} for every image in the dataset. 
    A cumulative error histogram for errors up to $0.25$ is generated and its area under the curve (AUC) determined for a set of images. 
    This horizon error AUC value gauges the overall accuracy of the estimated horizon lines. 
    We also report the mean squared error (MSE), which is more sensitive to outliers than the AUC. 
    In addition, we compute the estimated camera pose vector $\vec{p} \propto \mat{R} \vec{g} \propto \mat{K}\tran \vec{h}$ via inversion of Eq.~\ref{eq:gravity_to_horizon}.
    We determine the angular error $\xi$ between $\vec{p}$ and the ground truth pose $\vec{\hat{p}}$ for every image and report the AUC of the cumulative error histogram for  $\xi \leq 5 ^{\circ}$:
        $\cos \xi = \frac{\vec{p}\tran\vec{\hat{p}}}{\left\lVert \vec{p} \right\rVert_2 \left\lVert \vec{\hat{p}} \right\rVert_2} \, .$
    For applications that rely on horizon lines estimated from a video stream, it is desirable for the estimations be accurate as well as stable. 
    We propose another metric to measure undesirable fluctuations that do not reflect actual changes of the horizon over time: the average total variation $A_{TV}$. 
    For a sequence $n$ of length $T_n$ of estimated horizons $\vec{h}_{n,t}$ and corresponding ground truth $\vec{\hat{h}}_{n,t}$, with $t \in [1,T_n]$ and $n \in [1,N]$, 
    we compute the derivative $\sfrac{\partial L_e^{n,t}}{\partial t}$ of the horizon error according to Eq.~\ref{eq:max_horizon_error} using second order approximation.
    With $M = \sum_{n=1}^{N} T_n$ being the total number of images, the mean of its absolute calculated over all sequences yields the average total variation:\\
    \begin{equation}
        A_{TV} = \frac{1}{M} \sum_{n=1}^{N} \sum_{t=1}^{T_n} \left|\frac{\partial L_e^{n,t}}{\partial t}\right| \, .
    \end{equation}
    This metric is invariant to constant deviations from the ground truth but sensitive to higher frequency fluctuations.
    \begin{figure}	
		\centering
		\begin{subfigure}[t]{0.45\textwidth}
			\centering
			\includegraphics[width=0.99\linewidth]{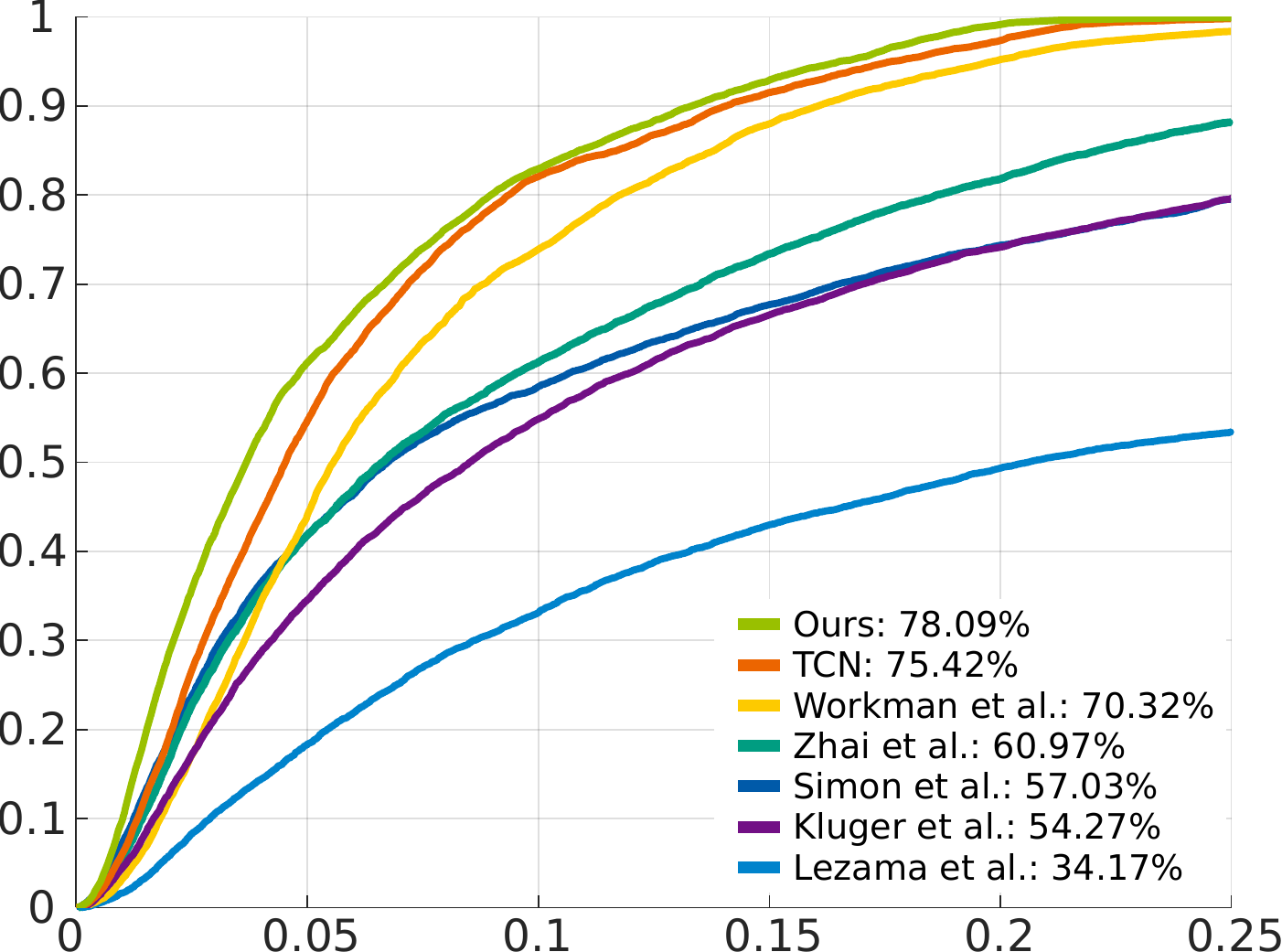}
			\caption{validation set}\label{fig:error_hist_val}		
		\end{subfigure}
		\quad
		\begin{subfigure}[t]{0.45\textwidth}
			\centering
			\includegraphics[width=0.99\linewidth]{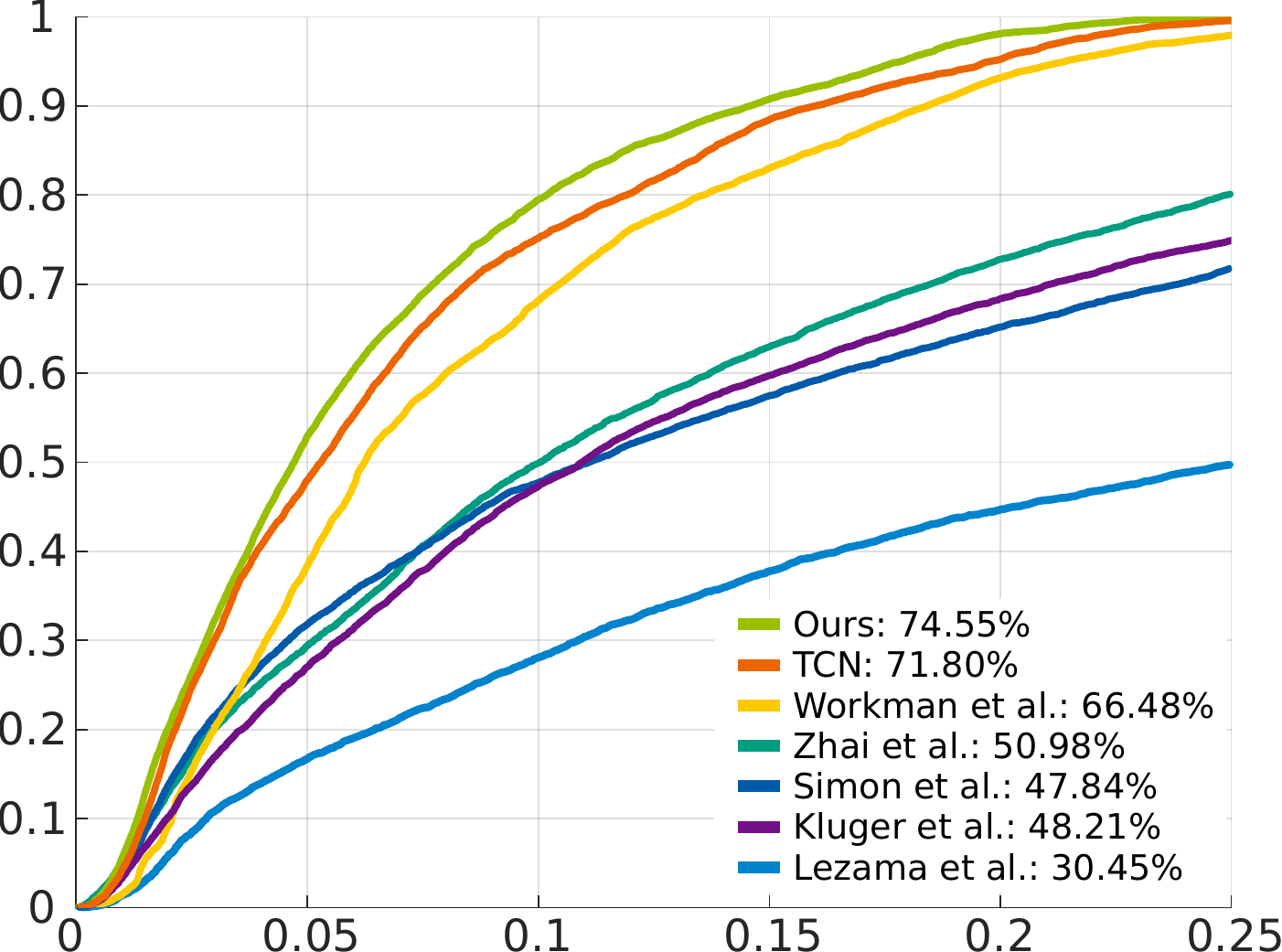}
			\caption{test set}\label{fig:error_hist_test}
		\end{subfigure}
		\caption{Cumulative horizon error histograms with AUC values for KITTI.}
		\label{fig:error_hist}
	\end{figure}   
	
    \subsection{KITTI Horizon Results}
    \label{sec:kitti_horizon_results}
    We report all metrics on the KITTI Horizon validation and test set for for following single frame algorithms: the VP based methods of Lezama \etal~\cite{lezama2014finding}, Kluger \etal~\cite{kluger2017deep} and Simon \etal~\cite{Simon_2018_ECCV}, the hybrid approach of Zhai \etal~\cite{zhai2016detecting} and the CNN based approach of Workman \etal~\cite{workman2016hlw}. 
    We also include results for our single frame CNN baseline (cf. Sec.~\ref{sec:single_image_estimation}), trained on either HLW or KITTI, for an \emph{average} baseline which simply always predicts the mean of the training set, a TCN~\cite{bai2018empirical} based temporally consistent approach with causal convolutions in the last three layers, and of course for our temporally consistent pipeline presented in Sec.~\ref{sec:temporally_consistent_estimation}. 
    The results are listed in Tab.~\ref{tab:kitti_results} and Fig.~\ref{fig:error_hist}.
    As these numbers show, methods based on line segments and vanishing points~\cite{kluger2017deep,lezama2014finding,Simon_2018_ECCV,zhai2016detecting} are unable to deliver consistent and accurate horizon estimates on KITTI. The best performing method among them is Zhai \etal~\cite{zhai2016detecting} with 60.97\%/50.98\% AUC (validation/test), which still lags behind the simplest \emph{average} baseline (69.40\%/64.18\%). 
    In addition, the very large mean squared error (MSE) and average total variation ($A_{TV}$) values -- up to several thousand -- indicate that these methods may fail catastrophically in some outlier cases. 
    In comparison, all CNN based methods -- including Workman \etal~\cite{workman2016hlw} and our own single frame CNN -- are significantly more accurate with at least 70.32\%/63.64\% AUC. 
    More importantly, the comparatively smaller MSE and $A_{TV}$ show that these methods are much less prone to extreme outliers. 
    If we compare the CNN of~\cite{workman2016hlw} with our own single-frame CNN trained on HLW, we observe that \cite{workman2016hlw} performs better overall -- all metrics but validation AUC are better to a relevant degree. 
    This is unsurprising, as~\cite{workman2016hlw} augmented their training with an additional 500000 images sampled from Google Street View, while we just used  HLW. 
    Naturally, if trained on the KITTI Horizon dataset, the accuracy of our single frame CNN increases significantly: from 71.10\%/63.64\% to 77.42\%/74.08\%, which is a 21.8\%/28.7\% relative increase. 
    Best results on all metrics are obtained with our temporally consistent approach (Sec.~\ref{sec:temporally_consistent_estimation}), with relative improvements upon the single frame CNN between 1.8\% (test AUC) and 12.1\% (test $A_{TV}$). 
    While the smoothness $A_{TV}$ of the single frame CNN improves measurably without diminishing overall accuracy if we additionally apply an exponential smoothing filter (Eq.~\ref{eq:exp_smooth}, $\alpha=0.5$), similar gains can be achieved when applied to the temporally consistent CNN as well, so it still retains its advantage. 
    We also trained a TCN~\cite{bai2018empirical} based on our single frame CNN with causal temporal convolutions of widths 3, 3, and 5 in the last three layers and a receptive field of nine frames. 
    Surprisingly, it performs worse than the single frame CNN on all metrics but $A_{TV}$. 
    We suspect that the TCN is more susceptible to overfitting, as it achieved a lower training loss, but higher validation loss compared to our other CNNs. 
    Compared to our ConvLSTM based network, it is on par \wrt $A_{TV}$ on the test set, but measurably worse otherwise.

	\subsection{Ablation Studies}
	\label{sec:ablation_study}
	\begin{table}
		\begin{center}
			\begin{tabular}{|l|c|c|c|}
				\cline{2-4}
				\multicolumn{1}{c|}{} &
				AUC & 
				MSE& 
				$A_{TV}$ \\ 
				\hline
				Huber loss (Sec.~\ref{sec:ablation_loss})&          71.96\%& 7.851 & 7.051\\
				\hline
				non-temporal (Sec.~\ref{sec:ablation_temporal})&    74.36\%& 7.266 & 5.699\\
				\hline
				w/o residual (Sec.~\ref{sec:ablation_convlstm})&    64.29\%& 11.60 & 5.279\\
				\hline
				na\"ive residual (Sec.~\ref{sec:ablation_convlstm})&74.01\%& 7.009 & \textbf{4.967}\\
				\hline
				Ours (Sec.~\ref{sec:temporally_consistent_estimation})&    \textbf{74.55}\%& \textbf{6.731} & {4.984} \\
				\hline
			\end{tabular}
		\end{center}
		\caption{Ablation study (Sec.~\ref{sec:ablation_study}) results on the KITTI Horizon test set. MSE and $A_{TV}$ scaled by $10^{-3}$.}
		\label{tab:ablation_results}
	\end{table}		
	
	\subsubsection{Loss function}
	\label{sec:ablation_loss}
	In order to investigate whether our new loss defined in Sec.~\ref{sec:loss_function} had the desired effect on estimation accuracy, we also trained our main CNN model described in Sec.~\ref{sec:horizon_line_network} using just the Huber loss defined in Eq.~\ref{eq:omega_theta_loss} and also used by~\cite{workman2016hlw}. 
	As Tab.~\ref{tab:ablation_results} shows, we report an AUC of 71.96\% and an MSE of $7.851 \cdot 10^{-3}$ on the test set. 
	Using our newly defined loss, however, we achieve an AUC of 74.55\% and an MSE of $6.731 \cdot 10^{-3}$, which marks a considerable relative improvement of 9.2\% and 14.3\% respectively.

    \subsubsection{Temporal information}
	\label{sec:ablation_temporal}
    As Tab.~\ref{tab:kitti_results} shows, our temporally consistent approach based on ConvLSTMs is able to achieve more accurate horizon estimates with significantly less variance. 
    In order to ascertain that this is due to the ConvLSTM utilising temporal correlations, and not simply due other architecture changes that arose as a result, we retrained our main CNN model with all temporal connections disabled, \ie we reset the LSTM states at every time step. 
    On the test set, this yields an AUC of 74.36\% and $A_{TV}$ of $5.699 \cdot 10^{-3}$. 
    When we enable the temporal connections of the LSTM, overall accuracy increases moderately -- 74.55\% AUC -- and $A_{TV}$ decreases noticeably to $4.984 \cdot 10^{-3}$, which is a relative improvement of 12.6\%. 
    We conclude that the ConvLSTM is indeed able to retain temporal consistency in a meaningful way.
	
	\subsubsection{ConvLSTM Architecture}
	\label{sec:ablation_convlstm}
	We compare our ConvLSTM architecture described in Sec~\ref{sec:residual_convlstm} against a ConvLSTM using a na\"ive residual path implementation and ConvLSTM without the residual path. 
	As Tab.~\ref{tab:ablation_results} shows, the na\"ive residual path already increases accuracy dramatically, from 64.29\% to 74.01\% AUC, and is evidently crucial for deep LSTM networks. 
	On par \wrt $A_{TV}$, our proposed ConvLSTM improves AUC and MSE upon the na\"ive implementation, yielding a relative improvement of 2.1\% and 4.0\% respectively. 
	While both approaches are able to generate smooth trajectories, our improved ConvLSTM is measurably more accurate on average.
	
	\section{Conclusion}
	The horizon line is an important geometric feature which can be used in many computer vision tasks, such as camera pose and ground plane estimation. %
	Due to their importance, horizon lines have received considerable attention in recent years. 
	Nonetheless, neither has any other work has focused on temporal consistency, nor are there appropriate datasets available.  
	In this work, an extension of the well-known KITTI database is presented that adds horizon line annotations to 72 sequences. 
	We furthermore propose a neural network for temporally consistent horizon line estimation in video sequences. 
	It utilises an improved convolutional LSTM and an adaptive loss function that yields more accurate horizon line estimates and ensures stable training. 
	The experimental evaluation demonstrates that the proposed architecture achieves superior performance for a diverse set of metrics which measure, for instance, accuracy and smoothness of trajectories.\\

    \paragraph{Acknowledgements} This work was supported by German Research Foundation (DFG) grant Ro 2497 / 12-2. \\

\renewcommand*\appendixpagename{\Large Appendix}

\begin{appendices}
	

This appendix is structured as follows: In Sec.~\ref{sec:supp_implementation_details}, we state more in-depth details of the implementation of our proposed method. 
Sec.~\ref{sec:supp_tcn} describes our implementation of Temporal Convolutional Networks, and provides additional experiments and evaluation. 
Sec.~\ref{sec:supp_results_quantitative} provides a more detailed look into the quantitative results of our proposed approach as well as its ablation studies, while Sec.~\ref{sec:supp_results_qualitative} contains horizon trajectories from KITTI Horizon to highlight a few best-, average- and worst-case examples of our approach.
In Sec.~\ref{sec:supp_kitti_dataset}, we provide a few examples from the KITTI Horizon dataset which convey an impression of the variety of scenes it contains. 
Sec.~\ref{sec:supp_hlw} discusses examples from the Horizon Line in the Wild dataset which have visibly inaccurate horizon line labels.

\section{Additional Implementation Details}
\label{sec:supp_implementation_details}
	We implemented all proposed neural network architectures using PyTorch~\cite{paszke2017automatic} version 0.4.1. 
	We used stochastic gradient descent with momentum ($0.9$) and $L_2$ regularisation ($10^{-4}$), and a cosine annealing learning rate schedule~\cite{loshchilov2016sgdr} starting at $10^{-1}$ and ending at $10^{-3}$. 
	On KITTI Horizon, all networks were trained for 160 epochs, and for 256 epochs on HLW. 
	Where applicable (see Sec.~\ref{sec:supp_results}), training was repeated four times with different random seeds, and the model with the highest validation AUC chosen. We downscale each image by a factor of two and apply the following augmentation techniques:
	\begin{itemize}
	    \item Random rotations $\beta \sim \mathcal{U}(-2^{\circ}, 2^{\circ})$  w.r.t. the image centre.
	    \item Random shifts $s_x \sim \mathcal{U}(-10~\mathrm{px}, 10~\mathrm{px})$ and $s_y \sim \mathcal{U}(-10~\mathrm{px}, 10~\mathrm{px})$.
	    \item Horizontal flips with probability $p=0.5$.
	    \item Colour jitter with brightness factor $\gamma_b \sim \mathcal{U}(0.75, 1.25)$, contrast factor $\gamma_c \sim \mathcal{U}(0.75, 1.25)$, saturation factor $\gamma_s \sim \mathcal{U}(0.75, 1.25)$ and hue factor $\gamma_h \sim \mathcal{U}(-0.25, 0.25)$.
	    \item Greyscale transformation with probability $p=0.1$.
	    \item Cutout~\cite{devries2017cutout} with width $w \sim \mathcal{U}(0, 512)$ and height $h \sim \mathcal{U}(0, 512)$.
	\end{itemize}
	We initialise the weights of the first nine convolutional layers of the networks from a ResNet18 pretrained on ImageNet~\cite{imagenet_cvpr09} while other layers are initialised randomly. 
	Training batches always contain $B$ sequences of $S$ consecutive frames from the KITTI Horizon training set, and batch size $B$ and sequence length $S$ were set to fulfil $S\cdot B = 128$. 
	We set $S=1$ for single frame approaches and $S=32$ for temporally consistent approaches. 
	Sampled sequences were always non-overlapping. At test time, we process the whole sequence as it appears in the dataset.

\section{Temporal Convolutional Networks}
\label{sec:supp_tcn}

	\begin{table*}
		\begin{center}
			\begin{tabular}{|l|cc|cc|cc|cc|}
				\cline{2-9}
				\multicolumn{1}{c}{}&
				\multicolumn{2}{|c}{AUC (horizon)} & 
				\multicolumn{2}{|c|}{MSE $\times 10^{-3}$ } & 
				\multicolumn{2}{|c|}{$A_{TV} \times 10^{-3}$ } &
				\multicolumn{2}{|c|}{AUC (pose) }\\
				\cline{1-9}
				\multicolumn{1}{|c|}{configuration} & 
				val & test &  
				val & test &  
				val & test &  
				val & test \\
				\hline
				1-3-3-5 ($S_{\mathrm{fov}}=9$)& 
                75.42\% & 71.80\%&
                6.392 & 8.318&
                4.946 & \textbf{4.937}&
                67.33\%& 64.21\%\\
				\hline
				1-3-5-5 ($S_{\mathrm{fov}}=11$)& 
                75.82\% & 71.65\%&
                6.498 & 8.329&
                4.997 & 5.119 &
                68.43\%& 64.14\%\\
				\hline
				1-3-5-7 ($S_{\mathrm{fov}}=13$)& 
                75.83\% & 72.23\%&
                6.383 & 7.909&
                4.932 & 5.043&
                68.42\%& 64.64\%\\
				\hline
				3-3-5-7 ($S_{\mathrm{fov}}=15$)& 
                76.08\% & 72.25\%&
                6.453 & 8.185&
                4.956 & 5.023&
                68.66\%& 64.69\%\\
				\hline
				3-5-5-7 ($S_{\mathrm{fov}}=17$)& 
                75.49\% & 71.57\%&
                7.084 & 8.458&
                5.117 & 5.263&
                68.52\%& 64.15\%\\
				\hline
				5-5-5-7 ($S_{\mathrm{fov}}=19$)& 
                75.76\% & 72.21\%&
                6.573 & 8.002&
                5.075 & 4.980&
                68.42\%& 64.73\%\\
				\hline
				\hline
				single frame&  
                77.42\% & 74.08\%&
                6.024 & 7.025&
                5.061 & 5.585 & 
                70.51\%& 66.62\%\\
				\hline
				\textbf{Ours} &
                \textbf{78.09\%}& \textbf{74.55\%}& 
                \textbf{5.427} & \textbf{6.731} &
                \textbf{4.619} & {4.984} & 
                \textbf{71.17\%}& \textbf{67.33\%}\\
                \hline
			\end{tabular}
		\end{center}
		\caption{Horizon estimation results on the KITTI Horizon validation and test sets comparing several TCN variants (Sec.~\ref{sec:supp_tcn}) with our single frame CNN and our proposed temporally consistent approach. }
		\label{tab:supp_tcn_results}
	\end{table*}	

\begin{figure*}	
		\centering
		\begin{subfigure}[t]{0.49\textwidth}
			\centering
			\includegraphics[width=\linewidth]{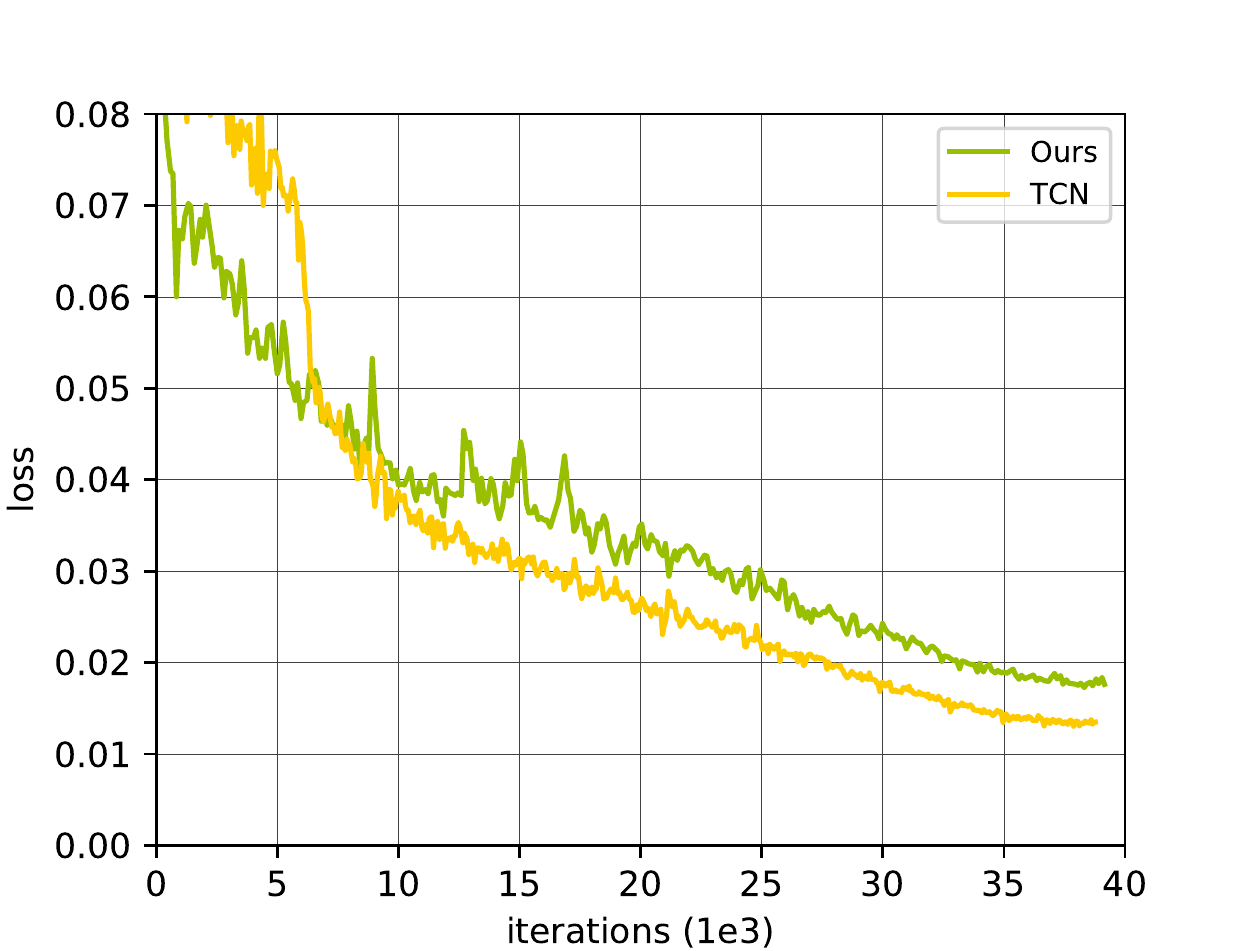}
			\caption{training loss}	
		\end{subfigure}
		\begin{subfigure}[t]{0.49\textwidth}
			\centering
			\includegraphics[width=\linewidth]{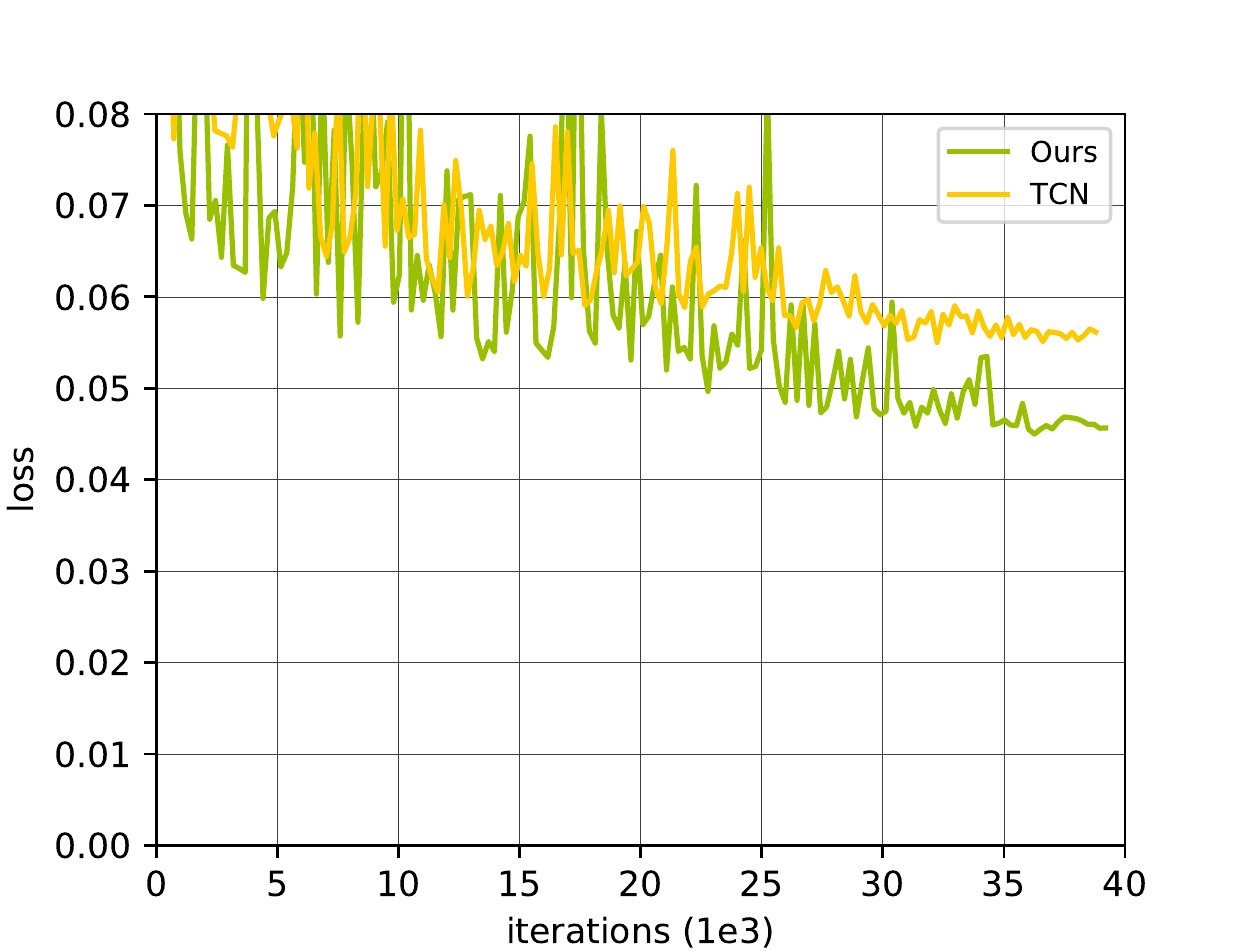}
			\caption{validation loss}
		\end{subfigure}
		\caption{Training and validation loss curves for our proposed ConvLSTM based CNN (\textcolor{nicegreen}{green}) and the 3-3-5-7 TCN (\textcolor{niceyellow}{yellow}, Sec.~\ref{sec:supp_tcn}).}
		\label{fig:supp_losses}
	\end{figure*}

As a possible alternative to our ConvLSTM based CNN, we briefly discussed Temporal Convolutional Networks (TCN)~\cite{bai2018empirical} in our paper. 
The authors of \cite{bai2018empirical} propose it as a purely feed-forward alternative to recurrent neural network structures -- such as LSTM and ConvLSTM -- for sequence modelling, and present promising results. 
We therefore implemented a TCN for the horizon line estimation task and compared it to our proposed ConvLSTM based architecture.
The concept of TCNs is based on causal convolutions along the temporal dimension of data. 
For a sequence of vectors $\vec{x}_t \in \mathbb{R}^C$, the 1D causal convolution across time with a kernel $\mat{h} \in \mathbb{R}^{M \times D \times C}$ can be defined as:
\begin{equation}
\vec{y}_t = \sum_{m=1}^{M} \vec{h}_m \vec{x}_{t-m+1} \, , \quad \vec{y}_t \in \mathbb{R}^D \, ,
\label{eq:supp_causal_convolution_1d}
\end{equation}
where $M$ denotes the number of elements of the sequence included in the convolution. 
Unlike a regular convolution, the result $\vec{y}_t$  of the causal convolution only depends on values of $\vec{x}_{\tau}$ for $\tau \leq t$, \ie no information from the future is considered. 
This can easily be generalised for sequences of images or feature maps $\vec{X}_t \in \mathbb{R}^{W \times H \times C}$ and a corresponding kernel $\mat{H} \in \mathbb{R}^{M \times A \times B \times D \times C}$:
\begin{equation}
\vec{Y}_{t} = \sum_{m=1}^{M} \vec{H}_{m} * \vec{X}_{t-m+1} \, ,
\label{eq:supp_causal_convolution_2d}
\end{equation}
where '$*$' denotes the 2D convolution operator commonly used in CNNs, $W$ and $H$ are image width and height, and $A \times B$ is the kernel size. 
Using regular convolutional layers readily available in deep learning frameworks, causal convolutional layers can be realised by simply shifting the output along the temporal axis by $\floor*{\sfrac{M}{2}}$ steps. 
If $L$ such layers with temporal convolution lengths $M_l$ are stacked to form a deeper network, the temporal field of view of this network becomes:
\begin{equation}
    S_{\mathrm{fov}} = 1-L+\sum_{l=1}^L M_l \, .
\end{equation}
We converted our ResNet18 based single-frame CNN into a TCN by replacing the last three respectively four 2D convolutional layers with 3D convolutional layers. 
We considered various configurations with $1 \leq M_l \leq 7$ and $9 \leq S_{\mathrm{fov}} \leq 19$, which we trained on the KITTI Horizon dataset as described in Sec.~\ref{sec:supp_implementation_details}. 
We named the configurations according to the values set for $M_l$ in the last four layers, \eg 1-3-3-5 means $M_{L-3:L} = [1,3,3,5]$. 
In order to avoid zero padding in the temporal dimension, we sampled additional $S_{\mathrm{fov}}-1$ previous frames for each sequence in a training batch, if possible. 
Tab.~\ref{tab:supp_tcn_results} shows the results of our TCNs on the KITTI Horizon validation and test sets. 
Compared to our ConvLSTM based approach, the TCNs perform poorly \wrt all metrics but $A_{TV}$. 
They even fall behind the single frame CNN \wrt AUC and MSE. We suspect that the TCNs are more susceptible to overfitting. 
In Fig.~\ref{fig:supp_losses}, we compare the training and validation losses of the 3-3-5-7 TCN with our proposed ConvLSTM network. 
The TCN achieves a noticeably lower training loss, but converges to a significantly higher validation loss. 
This indicates a lower ability of the TCN to generalise and may explain the poor validation and test performance.

\section{Additional Results}
\label{sec:supp_results}
\subsection{Quantitative}
\label{sec:supp_results_quantitative}
In order to gauge the uncertainty of our results arising from the randomness involved in neural network training, we repeated training for most of our experiments four times with varying random seeds. 
These experiments include: (a) our proposed ConvLSTM based CNN (\emph{Ours}), (b) the ablation study using the \emph{na\"ive residual} ConvLSTM, (c) the ablation study using the ConvLSTM with disabled temporal connections (\emph{non-temporal}) and (d) the \emph{single frame} CNN. 
In addition to the results for the training runs which perform best on the validation set, we also report mean and standard deviation over all four runs in Tab.~\ref{tab:supp_kitti_results}. 
As these results show, the single-frame and non-temporal variants perform worse than the temporally consistent approaches, even when averaged over multiple runs. 
Comparing our proposed ConvLSTM with the na\"ive residual variant, we observe similar performance \wrt $A_{TV}$ on both validation and test sets. 
AUC and MSE results on the validation set are similar as well, \wrt both best and mean performance. 
At the same time, the test performance of our proposed model with the best validation performance is measurably better than the na\"ive residual variant, which indicates an improved generalisation ability of our proposed ConvLSTM.

	\begin{table*}
		\begin{center}
			\begin{tabular}{|l|cc|cc|cc|}
				\cline{2-7}
				\multicolumn{1}{c}{}&
				\multicolumn{2}{|c}{AUC (horizon)} & 
				\multicolumn{2}{|c|}{MSE $\times 10^{-3}$ } & 
				\multicolumn{2}{|c|}{$A_{TV} \times 10^{-3}$ } \\
				\cline{2-7}
				\multicolumn{1}{c|}{} & 
				val & test &  
				val & test &  
				val & test \\
				\hline
				{single }& 
                77.42\% & 74.08\%&
                6.024 & 7.025&
                5.061 & 5.585 \\
				frame& 
				\small (77.03 $\pm$ 0.296)& \small (74.19 $\pm$ 0.219)&
				\small (6.125 $\pm$ 0.0635)& \small (7.084 $\pm$ 0.0886)&
				\small (5.276 $\pm$ 0.237)& \small (5.670 $\pm$ 0.0893)\\
				\hline
				non- & 
                77.63\% & 74.36\%&
                5.852 & 7.266&
                5.368 & 5.699 \\
				temporal& 
				\small (77.19 $\pm$ 0.514)& \small (74.43 $\pm$ 0.332)&
				\small (5.926 $\pm$ 0.154)& \small (7.122 $\pm$ 0.228)&
				\small (5.569 $\pm$ 0.234)& \small (5.982 $\pm$ 0.291)\\
				\hline
				na\"ive  & 
                \textbf{78.19\%} & 74.01\%&
                5.534 & 7.009&
                \textbf{4.583} & \textbf{4.967} \\
				residual& 
				\small (77.74 $\pm$ 0.298)& \small (74.19 $\pm$ 0.262)&
				\small (5.723 $\pm$ 0.120)& \small (6.980 $\pm$ 0.122)&
				\small (4.705 $\pm$ 0.212)& \small (5.056 $\pm$ 0.0988)\\
				\hline
					\multicolumn{1}{|l|}{\textbf{Ours}}& 
				{78.09\%}& \textbf{74.55\%}& 
                \textbf{5.427} & \textbf{6.731} &
                {4.619} & {4.984} \\
                &
				\small (77.60 $\pm$ 0.296)& \small (74.42 $\pm$ 0.233)&
				\small (5.760 $\pm$ 0.208)& \small (7.024 $\pm$ 0.206)&
				\small (4.716 $\pm$ 0.0912)& \small (5.071 $\pm$ 0.0687)\\
				\hline
			\end{tabular}
		\end{center}
		\caption{Horizon estimation results on the KITTI Horizon validation and test sets. We compare our proposed temporally consistent approach with two ablation studies and our single-frame CNN, see Sec.~\ref{sec:supp_results_quantitative}. We present the results of the training run with the best validation AUC out of four runs. The numbers in brackets are (mean $\pm$ standard deviation) over all four runs. }
		\label{tab:supp_kitti_results}
	\end{table*}	

\subsection{Qualitative}
\label{sec:supp_results_qualitative}
\begin{figure*}	
		\centering

		\begin{subfigure}[t]{\textwidth}
			\centering
			\begin{minipage}[t]{0.49\textwidth}
			\vspace{0pt}
			\centering
			\hspace{1.5em}
			\includegraphics[width=0.875\textwidth]{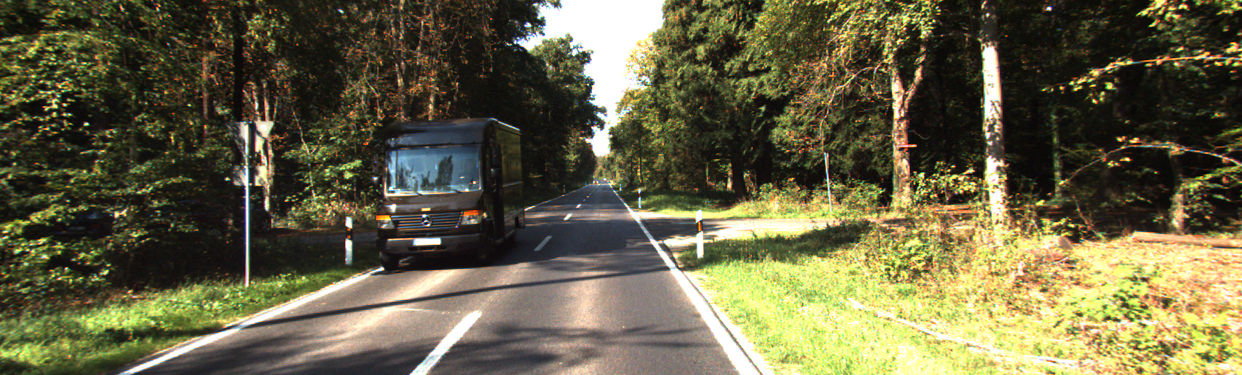}
			\end{minipage}
			\begin{minipage}[t]{0.49\textwidth}
			\vspace{0pt}
			\includegraphics[width=\textwidth]{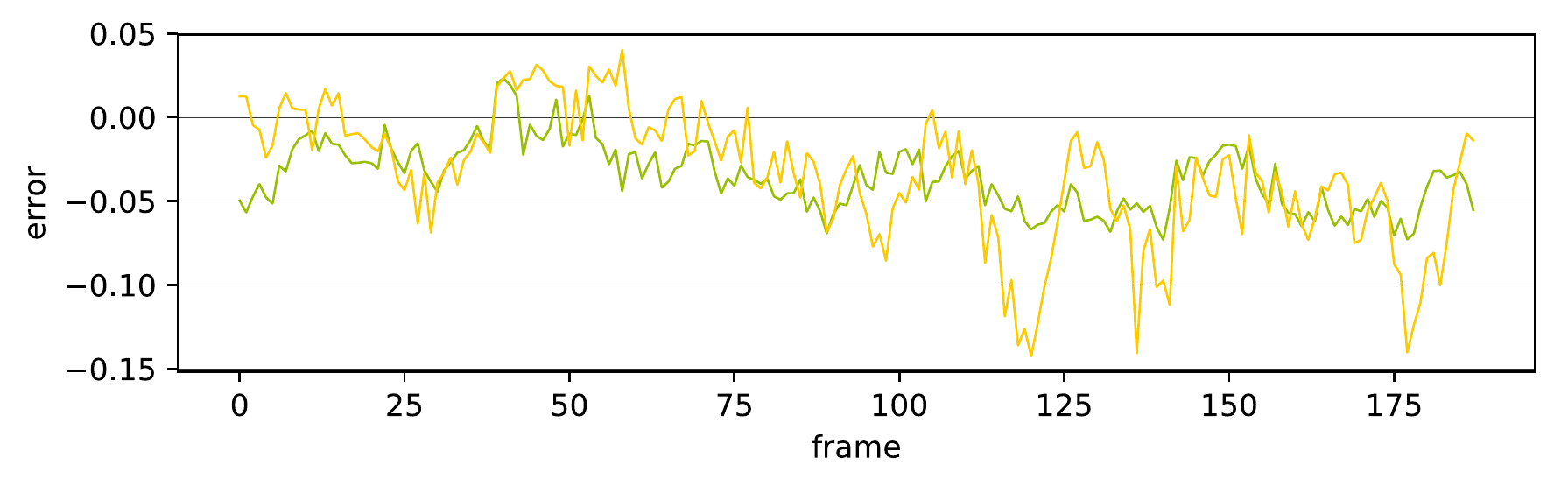}
			\end{minipage}
			\includegraphics[width=0.49\textwidth]{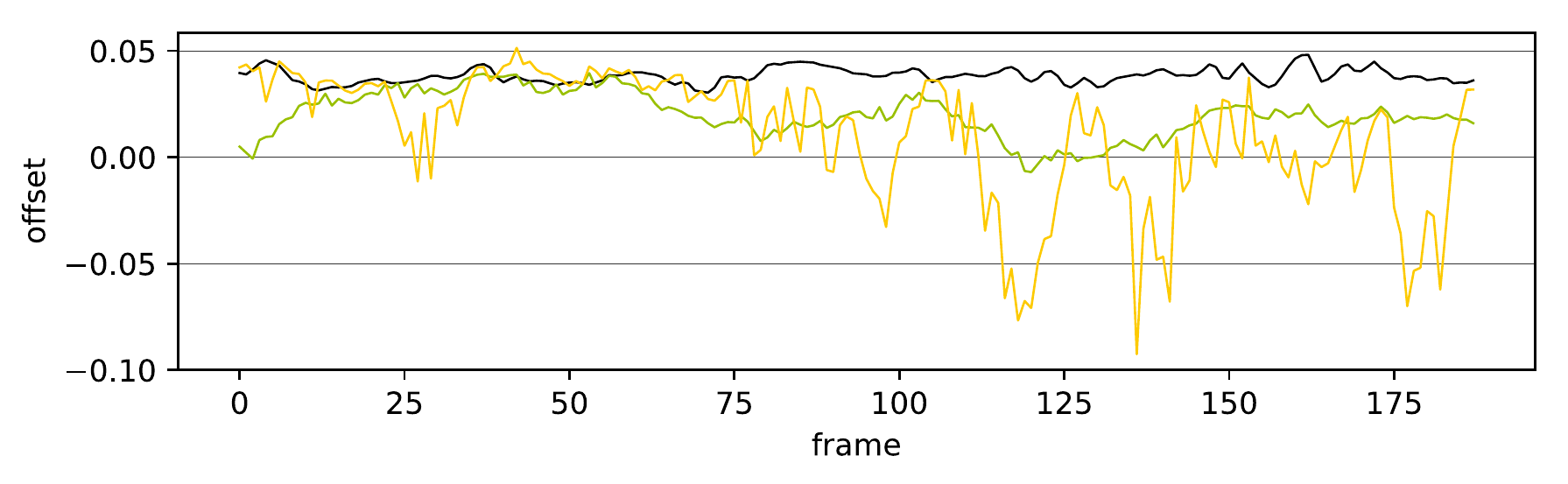}
			\includegraphics[width=0.49\textwidth]{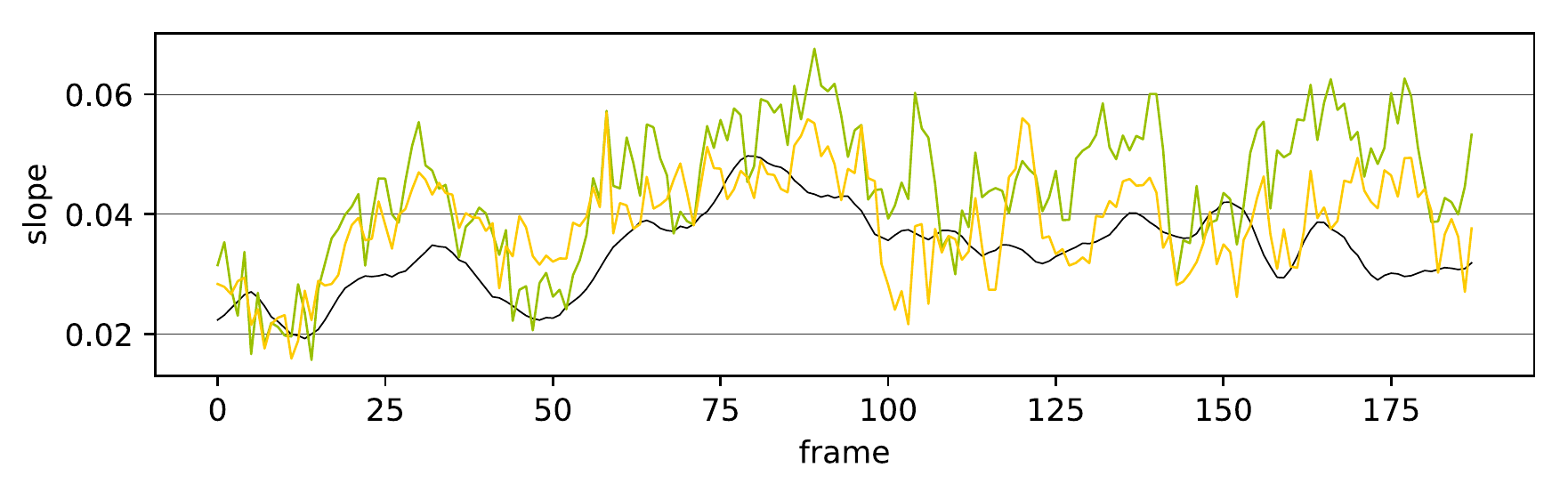}
			\caption{Example 1}
			\label{fig:supp_kitti_example_1}
		\end{subfigure}
		\vspace{1em}
		
		\begin{subfigure}[t]{\textwidth}
			\centering
			\begin{minipage}[t]{0.49\textwidth}
			\vspace{0pt}
			\centering
			\hspace{1.5em}
			\includegraphics[width=0.875\textwidth]{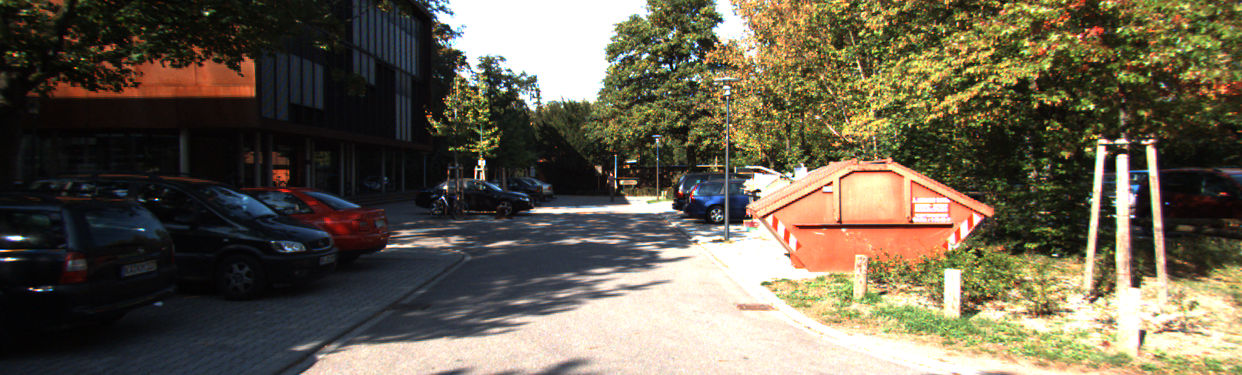}
			\end{minipage}
			\begin{minipage}[t]{0.49\textwidth}
			\vspace{0pt}
			\includegraphics[width=\textwidth]{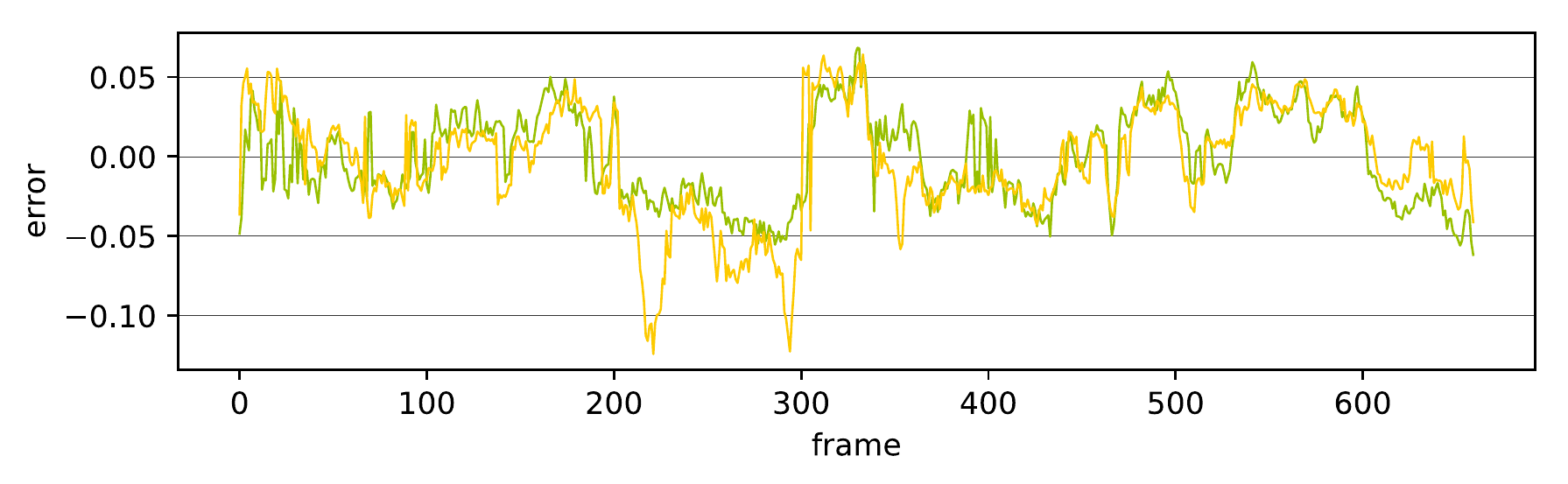}
			\end{minipage}
			\includegraphics[width=0.49\textwidth]{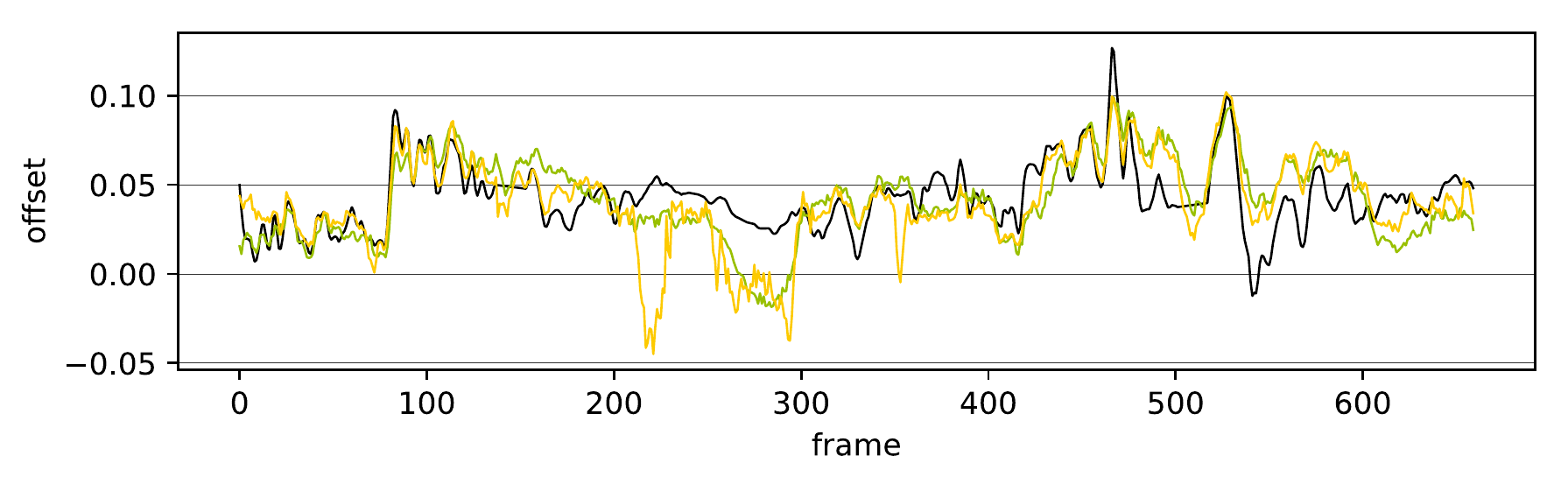}
			\includegraphics[width=0.49\textwidth]{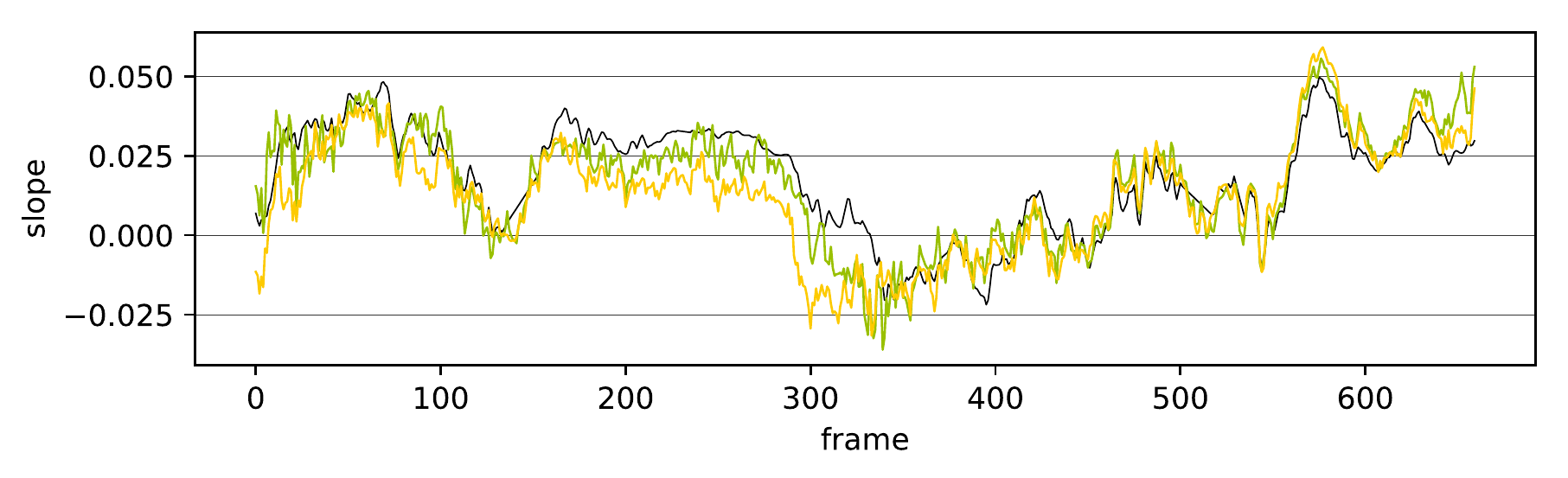}
			\caption{Example 2}
			\label{fig:supp_kitti_example_2}
		\end{subfigure}
		\vspace{1em}
		
		\begin{subfigure}[t]{\textwidth}
			\centering
			\begin{minipage}[t]{0.49\textwidth}
			\vspace{0pt}
			\centering
			\hspace{1.5em}
			\includegraphics[width=0.875\textwidth]{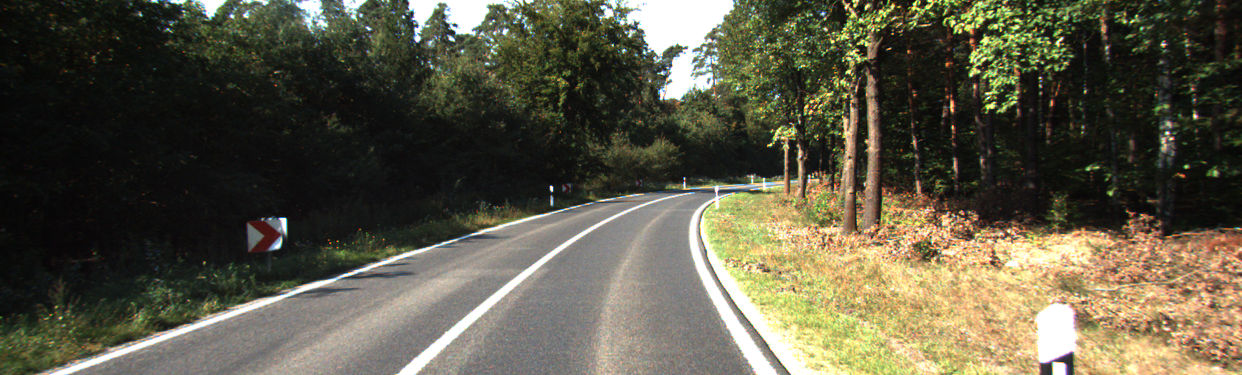}
			\end{minipage}
			\begin{minipage}[t]{0.49\textwidth}
			\vspace{0pt}
			\includegraphics[width=\textwidth]{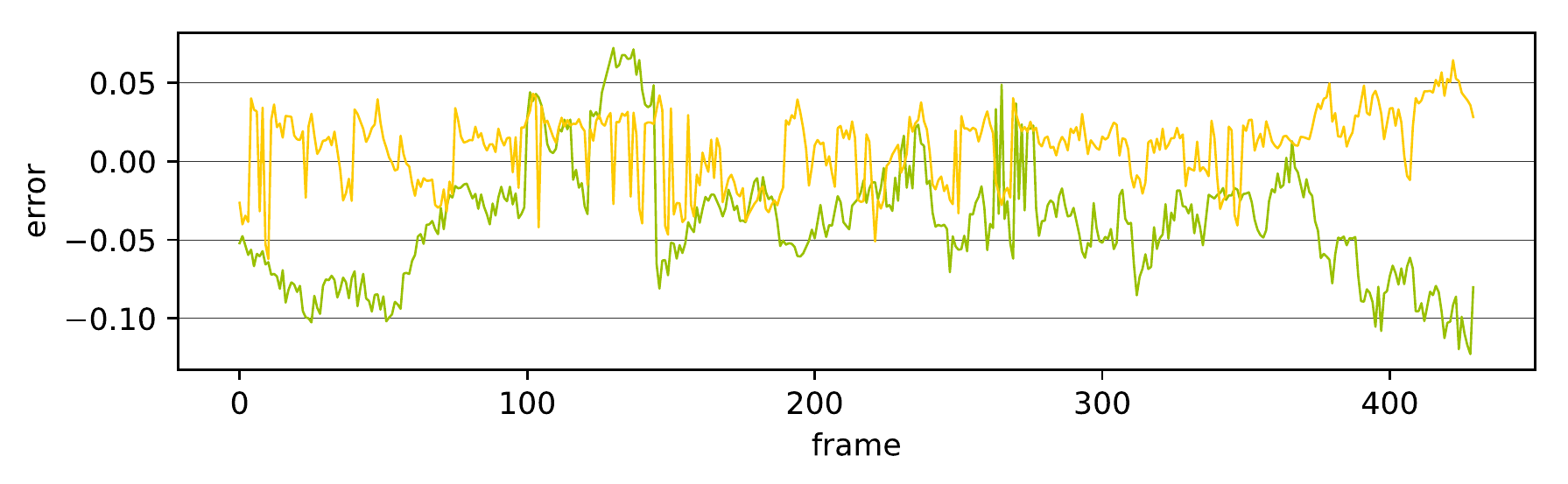}
			\end{minipage}
			\includegraphics[width=0.49\textwidth]{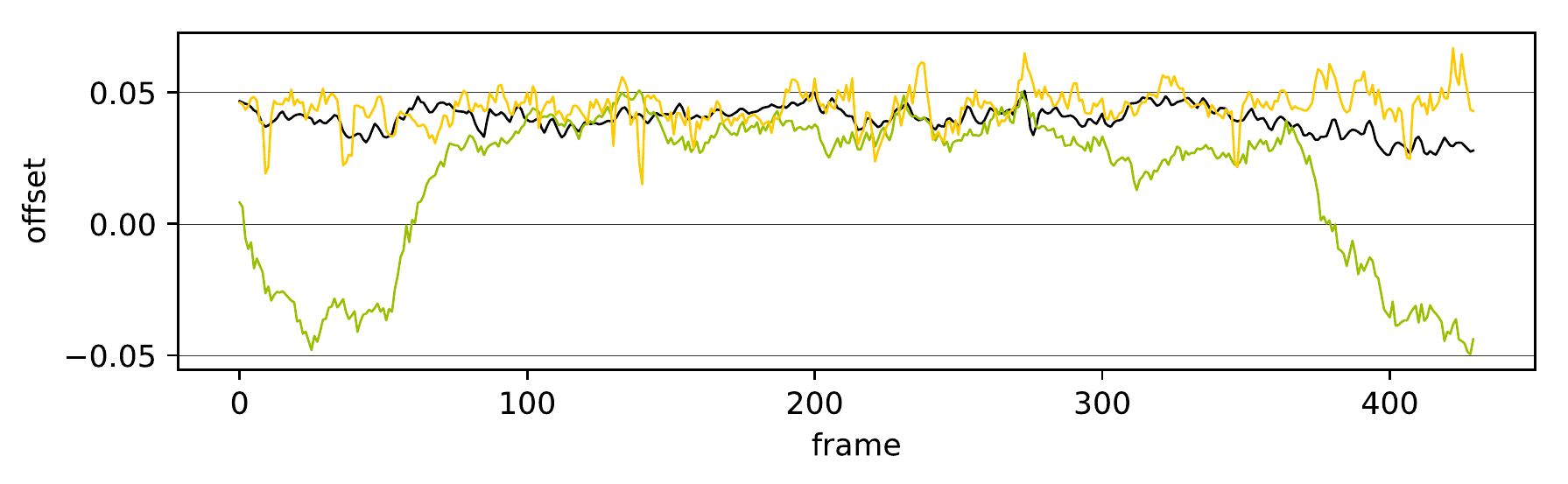}
			\includegraphics[width=0.49\textwidth]{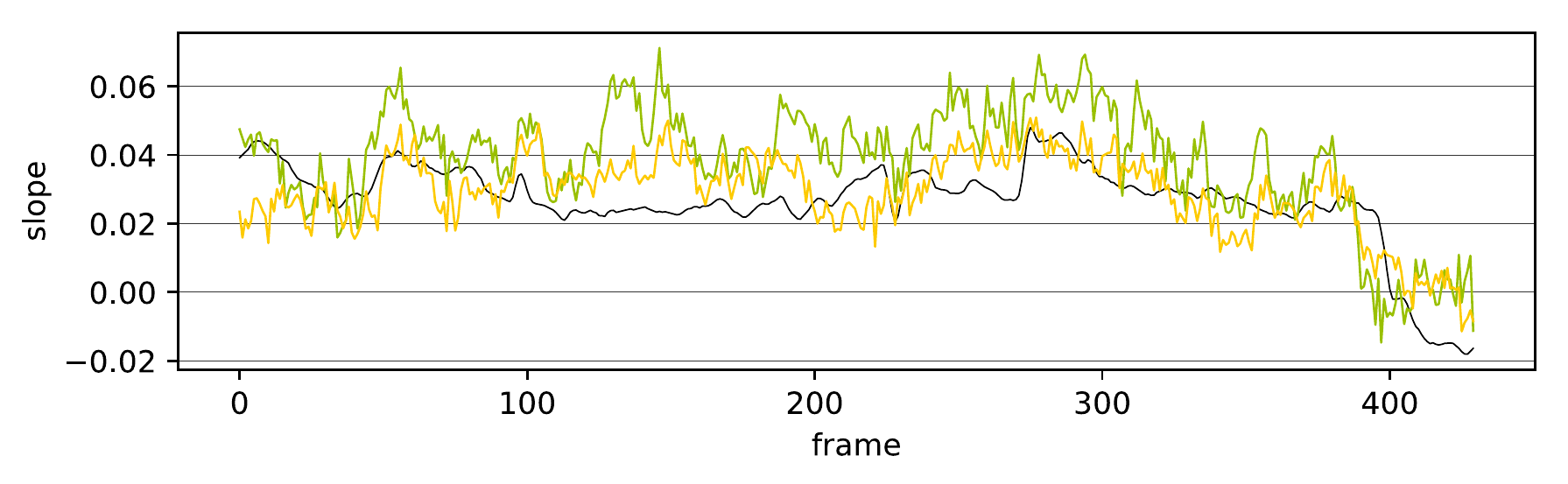}
			\caption{Example 3}
			\label{fig:supp_kitti_example_3}
		\end{subfigure}
		\vspace{1em}

		\caption{Three example horizon line trajectories from the KITTI Horizon dataset. We compare our proposed temporally consistent approach (\textcolor{nicegreen}{green}) with the single frame CNN (\textcolor{niceyellow}{yellow}) and the ground truth (black). Top left: example image from the video sequence. Top right: signed horizon error over time. Bottom: horizon offset and slope over time.}
		\label{fig:supp_kitti_examples}
	\end{figure*}

In Fig.~\ref{fig:supp_kitti_examples}, we show three example horizon line trajectories from the KITTI Horizon dataset. 
In the first example, Fig.~\ref{fig:supp_kitti_example_1}, the single frame estimation fluctuates heavily, while our proposed temporally consistent approach remains much more stable throughout the sequence. 
The second example, Fig.~\ref{fig:supp_kitti_example_2}, contains segments where the single frame estimation is moderately better, \eg between frames 150 and 200, but also segments where the single frame estimation shows severe fluctuations, \eg frames 200 to 300. 
Besides that, the results of the two algorithms are mostly very similar. 
Lastly, in Fig.~\ref{fig:supp_kitti_example_3}, we can observe a failure case of our approach. 
In the middle section between frames 100 and 300, both algorithms perform similarly, but at the beginning and at the end, our proposed method exhibits a relatively constant but large error. 

\subsection{KITTI Horizon Dataset}
\label{sec:supp_kitti_dataset}
\begin{figure*}	
		\centering
		\begin{subfigure}[t]{0.49\textwidth}
			\centering
			\includegraphics[width=\linewidth]{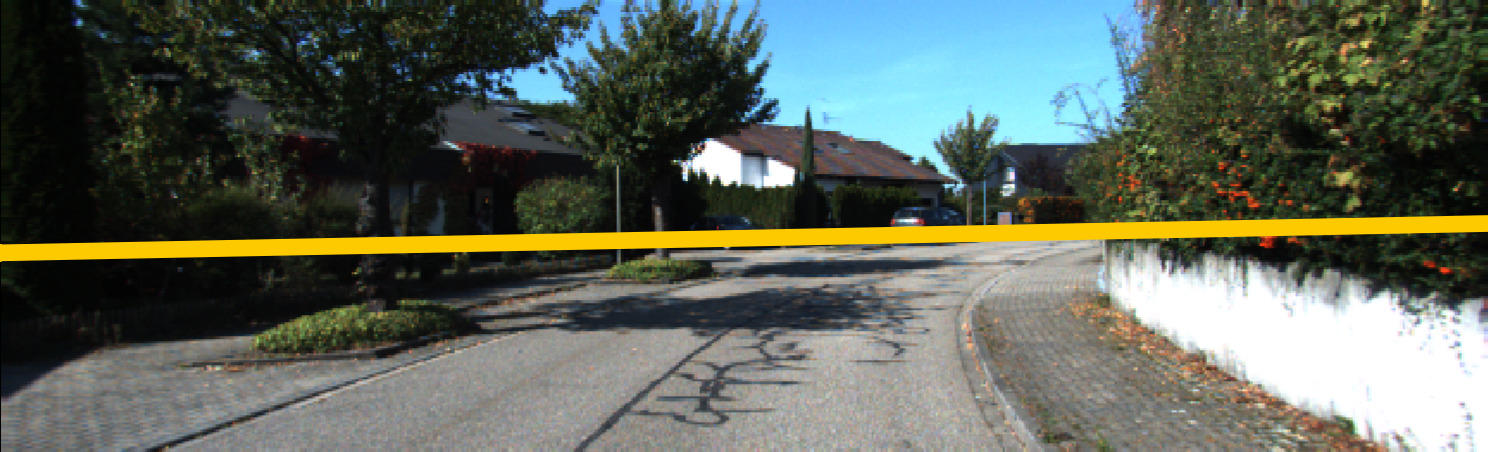}
		\end{subfigure}
		\begin{subfigure}[t]{0.49\textwidth}
			\centering
			\includegraphics[width=\linewidth]{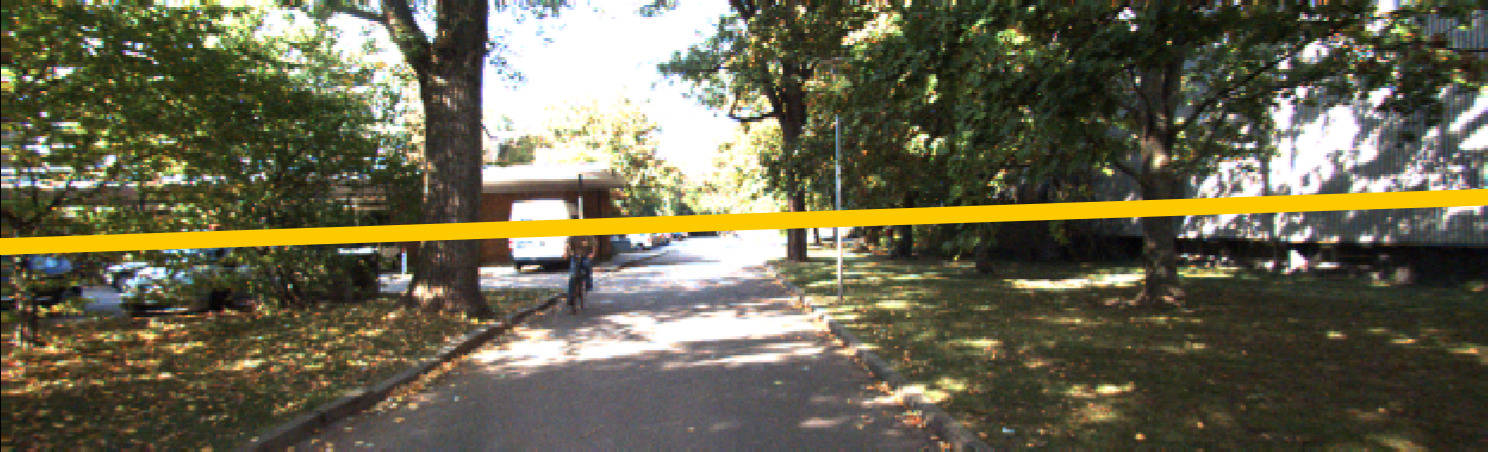}
		\end{subfigure}
		\begin{subfigure}[t]{0.49\textwidth}
			\centering
			\includegraphics[width=\linewidth]{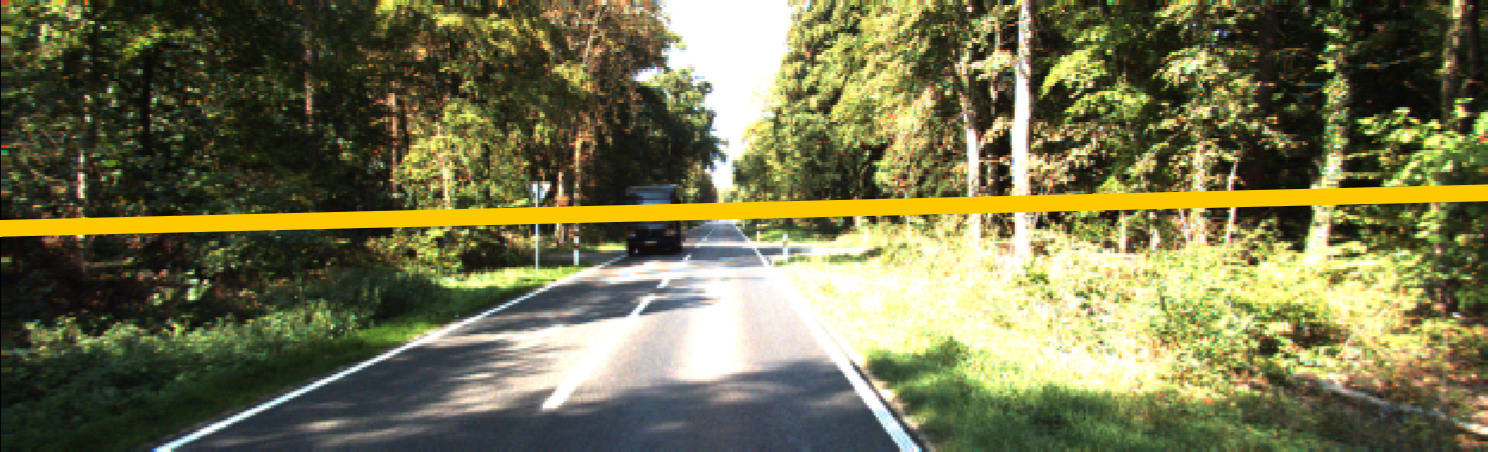}
		\end{subfigure}
		\begin{subfigure}[t]{0.49\textwidth}
			\centering
			\includegraphics[width=\linewidth]{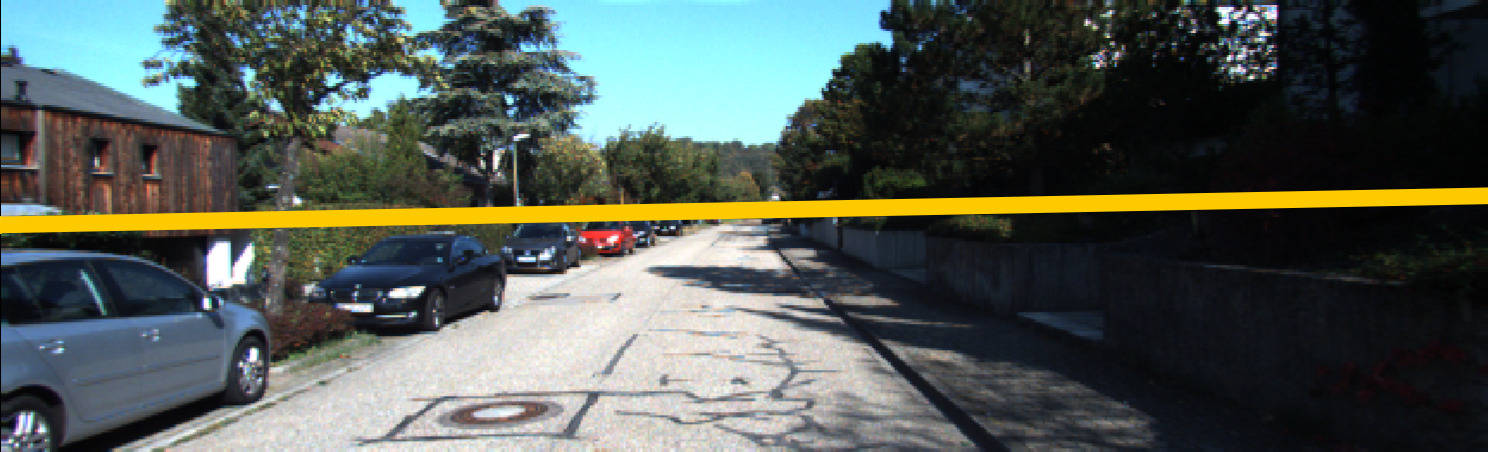}
		\end{subfigure}
		\begin{subfigure}[t]{0.49\textwidth}
			\centering
			\includegraphics[width=\linewidth]{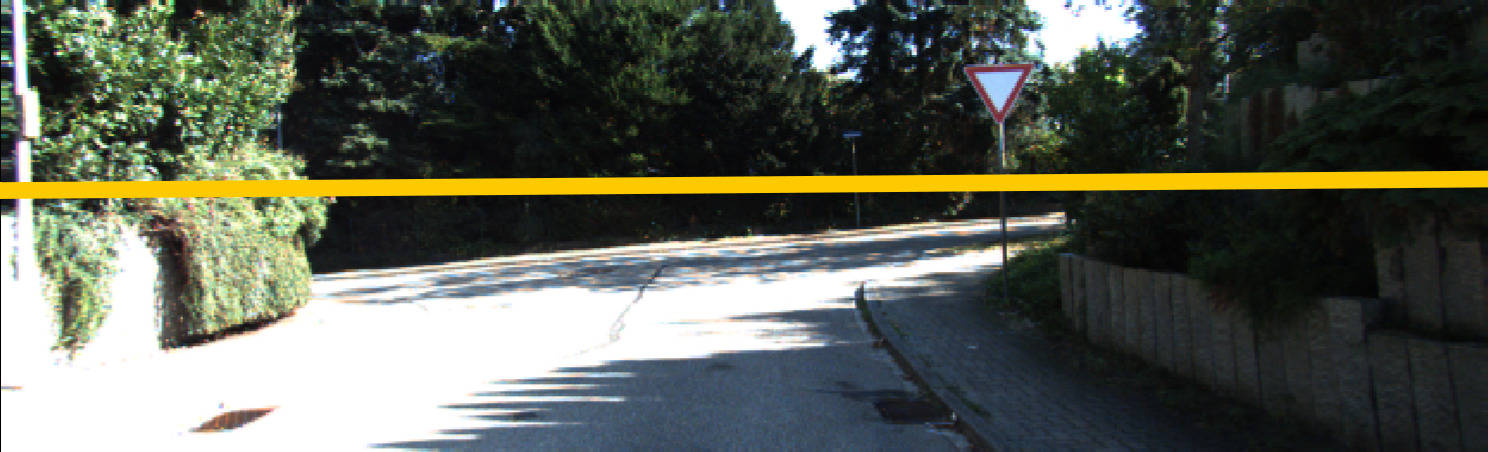}
		\end{subfigure}
		\begin{subfigure}[t]{0.49\textwidth}
			\centering
			\includegraphics[width=\linewidth]{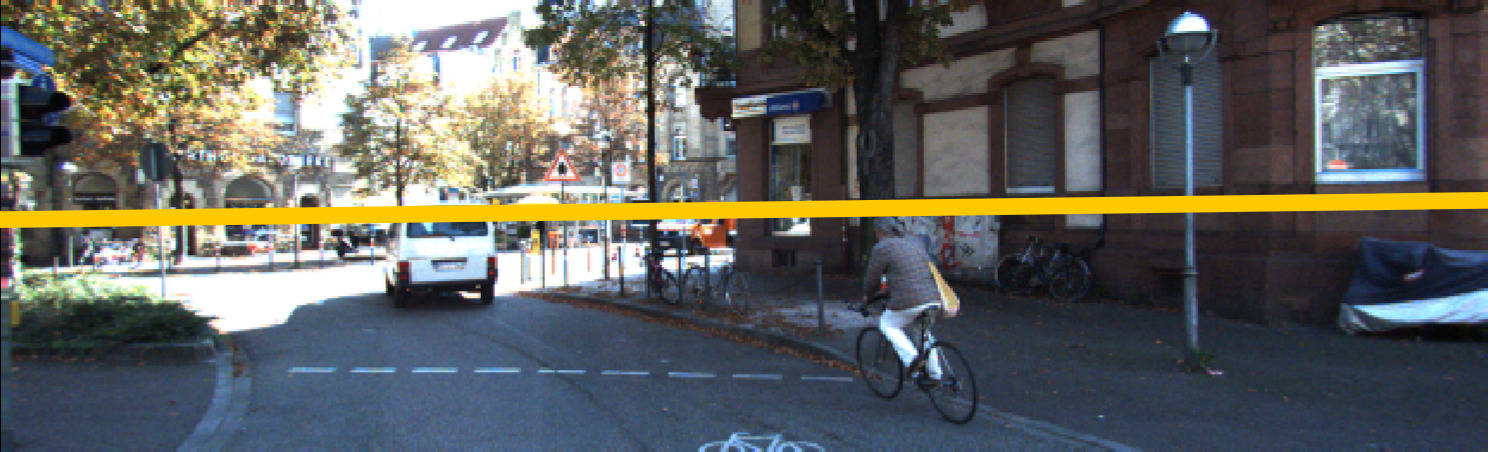}
		\end{subfigure}
		\begin{subfigure}[t]{0.49\textwidth}
			\centering
			\includegraphics[width=\linewidth]{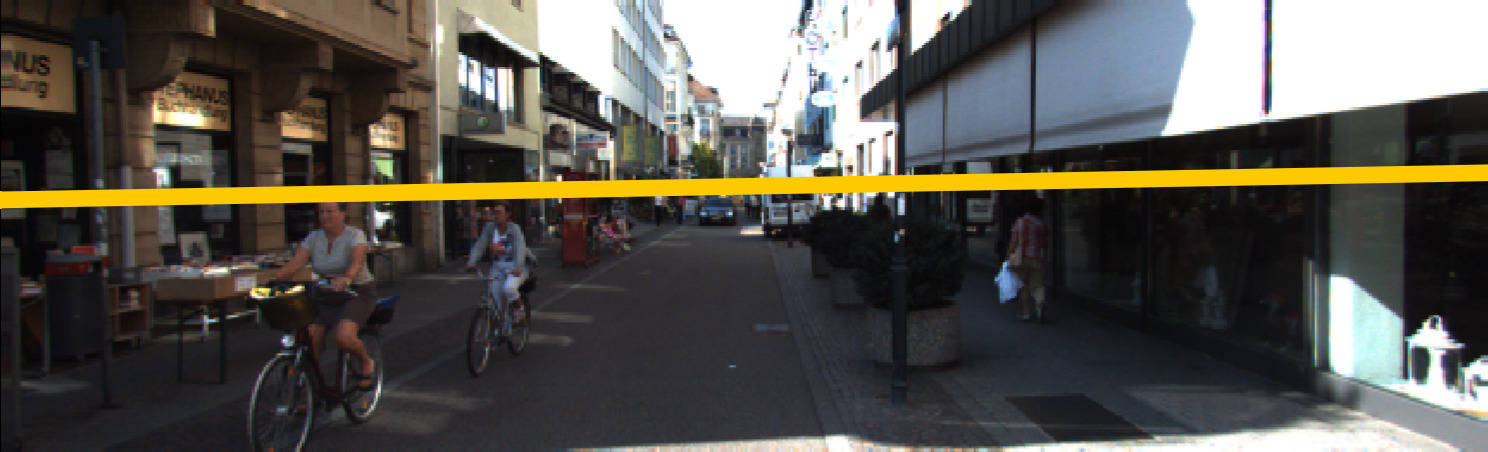}
		\end{subfigure}
		\begin{subfigure}[t]{0.49\textwidth}
			\centering
			\includegraphics[width=\linewidth]{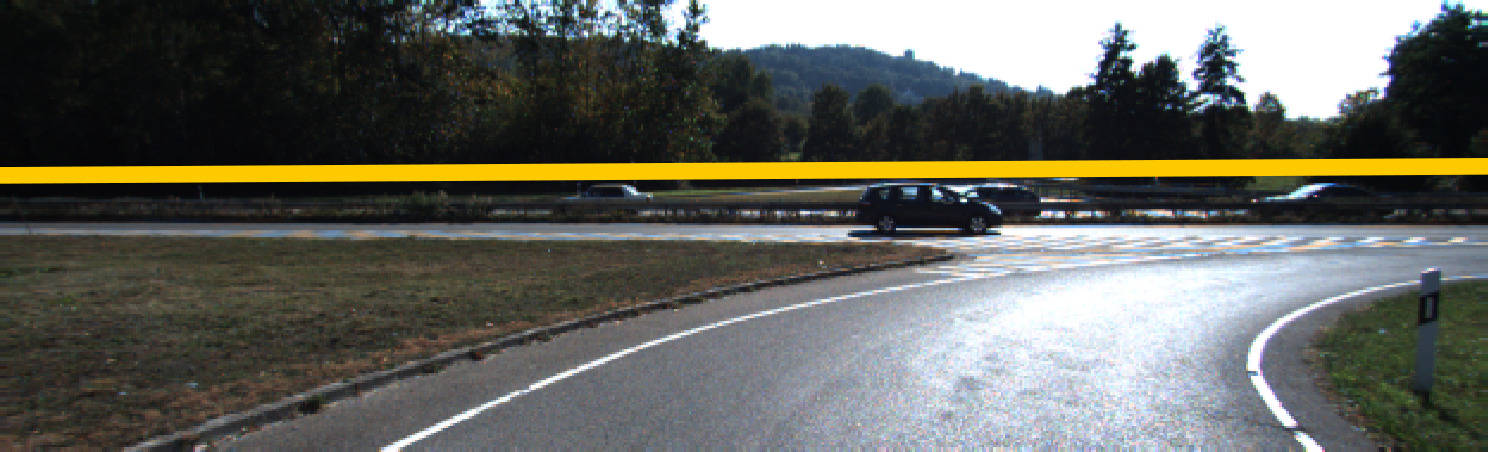}
		\end{subfigure}
		\begin{subfigure}[t]{0.49\textwidth}
			\centering
			\includegraphics[width=\linewidth]{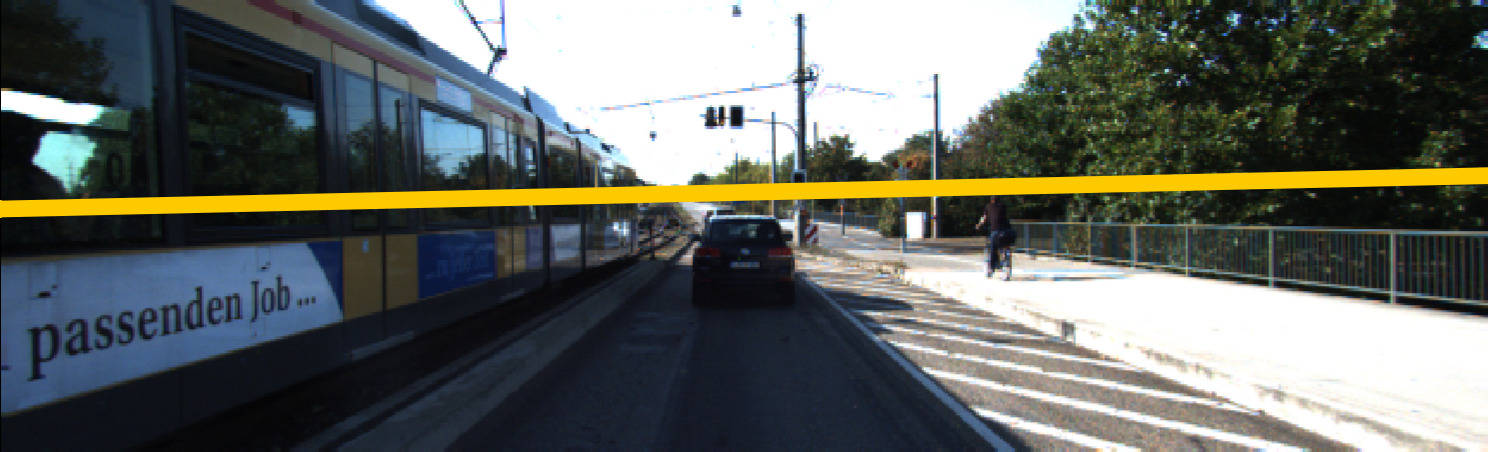}
		\end{subfigure}
		\begin{subfigure}[t]{0.49\textwidth}
			\centering
			\includegraphics[width=\linewidth]{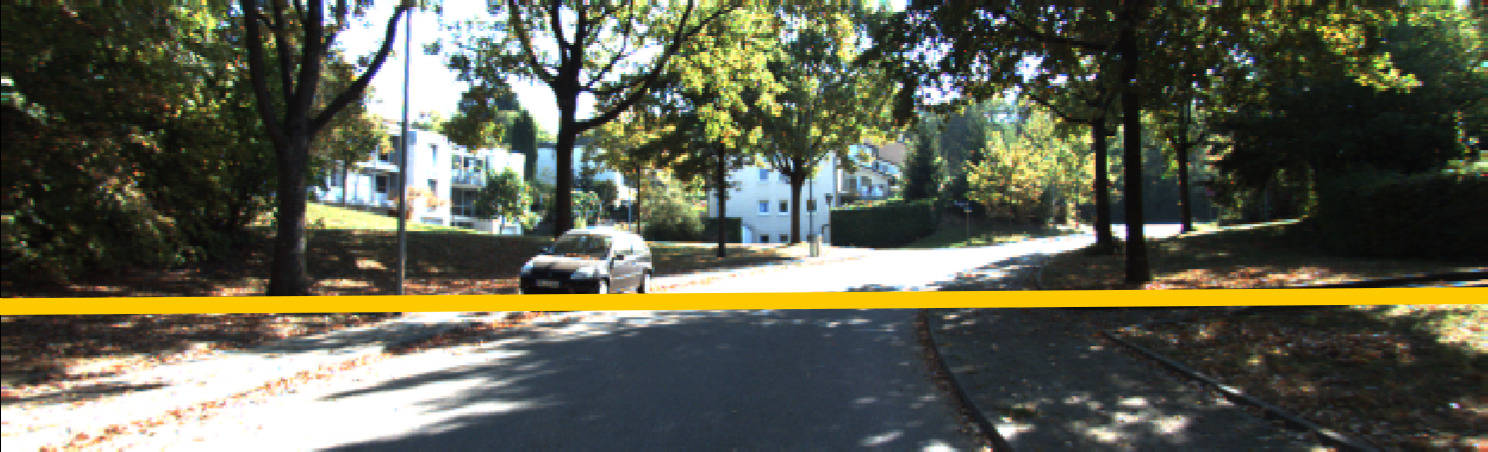}
		\end{subfigure}
		\begin{subfigure}[t]{0.49\textwidth}
			\centering
			\includegraphics[width=\linewidth]{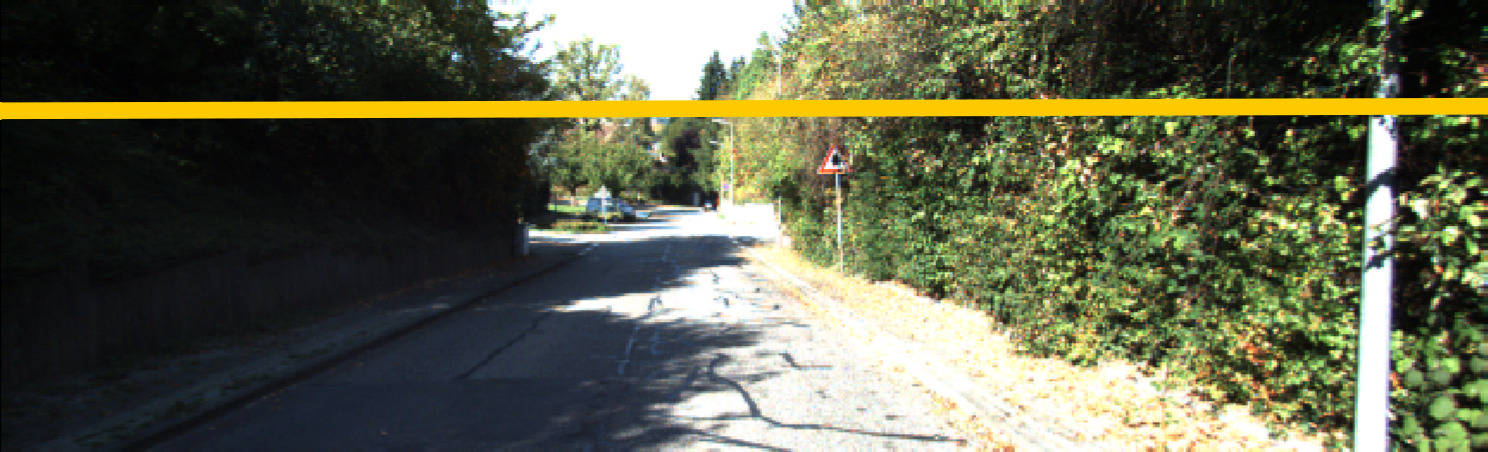}
		\end{subfigure}
		\begin{subfigure}[t]{0.49\textwidth}
			\centering
			\includegraphics[width=\linewidth]{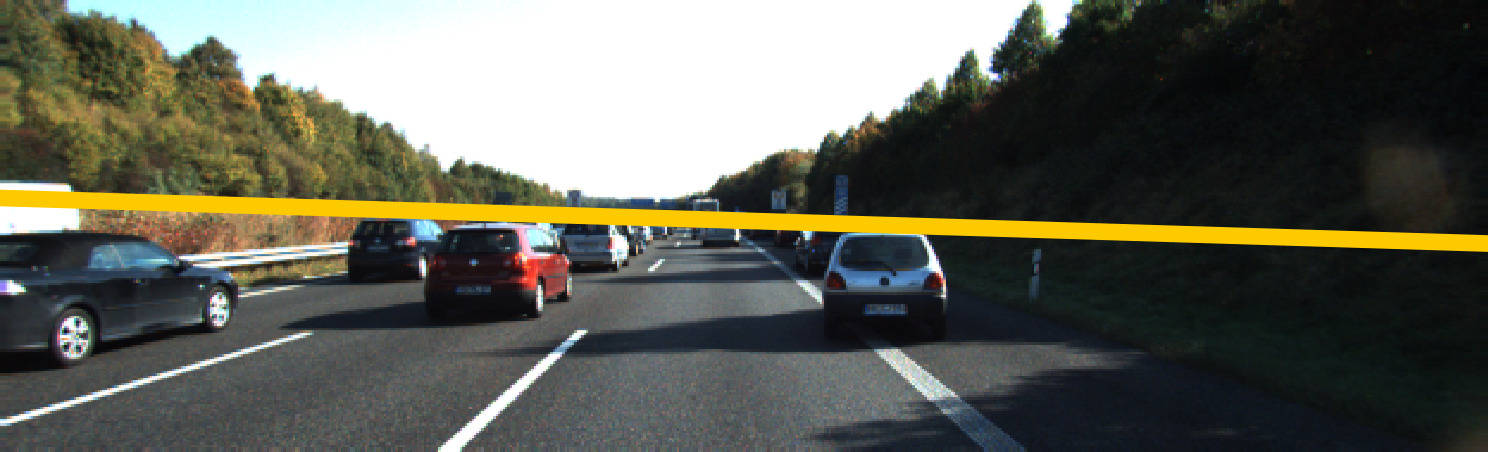}
		\end{subfigure}
		\begin{subfigure}[t]{0.49\textwidth}
			\centering
			\includegraphics[width=\linewidth]{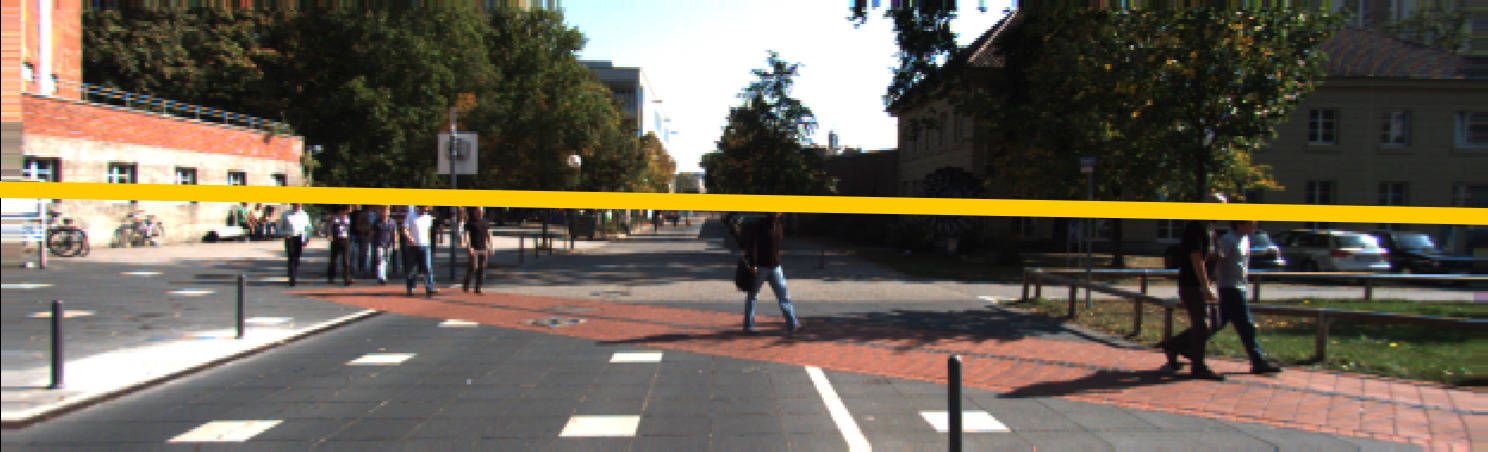}
		\end{subfigure}
		\begin{subfigure}[t]{0.49\textwidth}
			\centering
			\includegraphics[width=\linewidth]{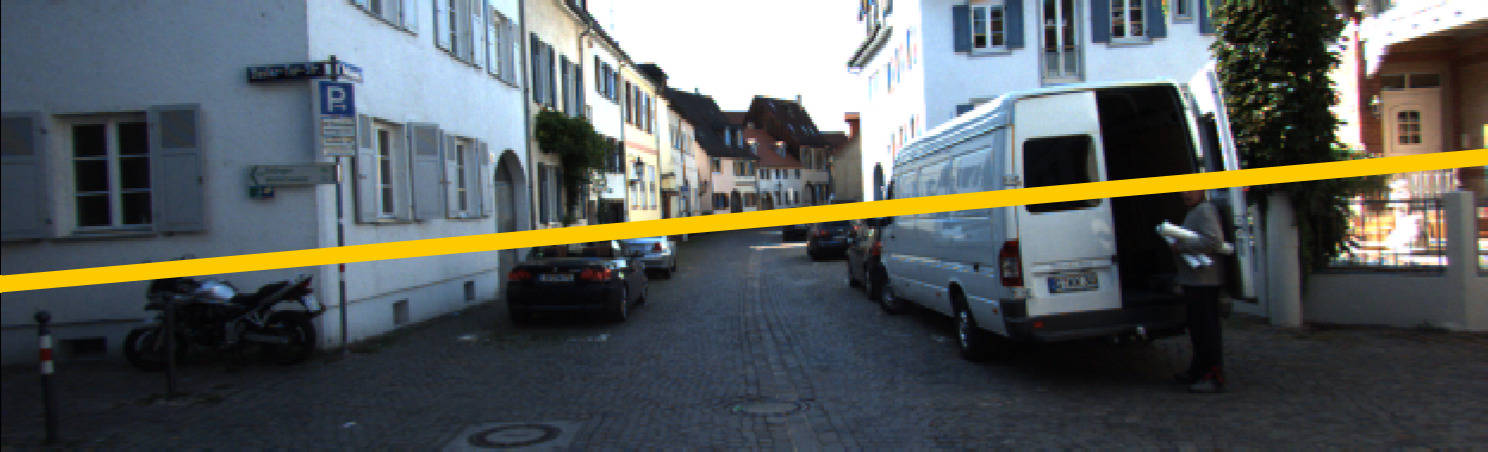}
		\end{subfigure}
		\begin{subfigure}[t]{0.49\textwidth}
			\centering
			\includegraphics[width=\linewidth]{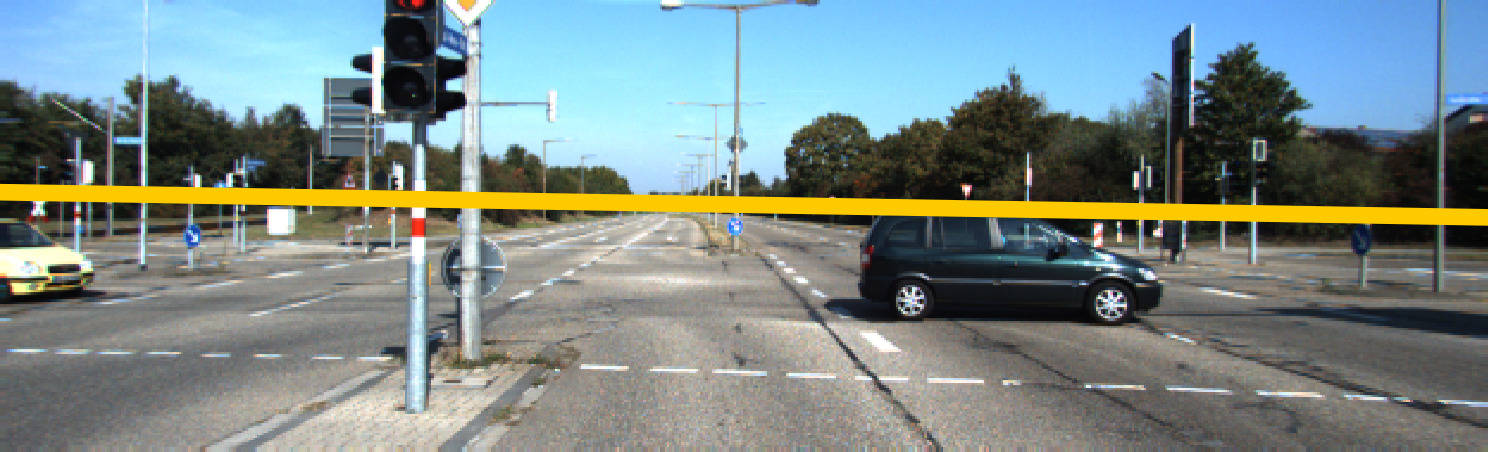}
		\end{subfigure}
		\begin{subfigure}[t]{0.49\textwidth}
			\centering
			\includegraphics[width=\linewidth]{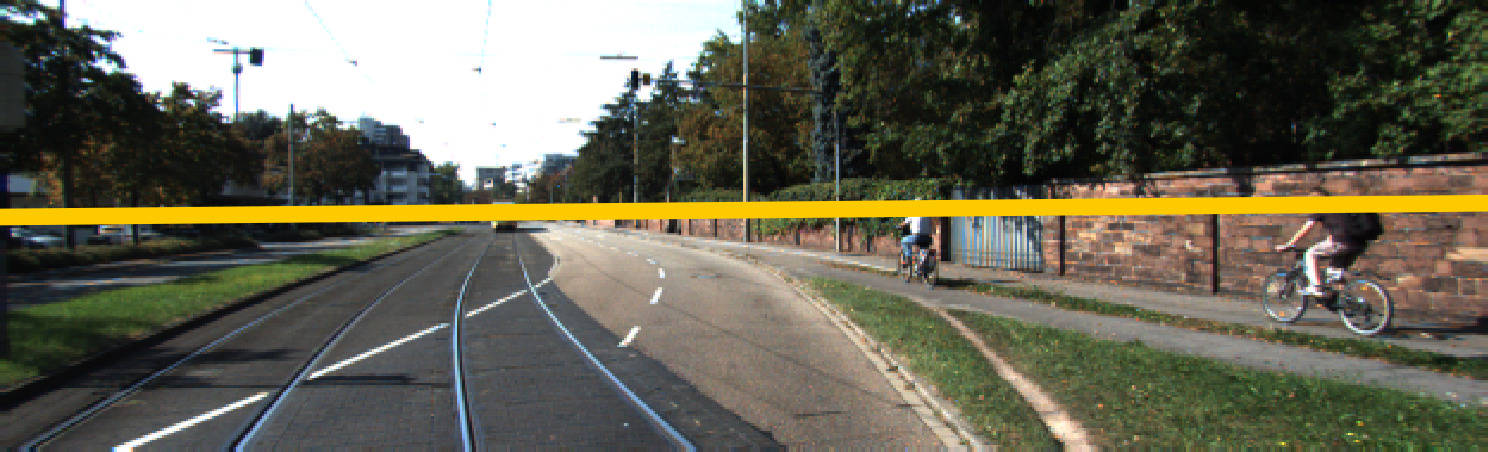}
		\end{subfigure}

		\caption{Examples from the KITTI Horizon dataset with ground truth horizon lines (\textcolor{niceyellow}{yellow}).}
		\label{fig:supp_kitti_gt_examples}
	\end{figure*}

We provide a few examples from the KITTI Horizon dataset with ground truth horizons in Fig.~\ref{fig:supp_kitti_gt_examples}, in order to give an impression of the variety of scenes it contains.

\section{Horizon Lines in the Wild}
\label{sec:supp_hlw}
\begin{figure*}	
		\centering
		\begin{subfigure}[t]{0.245\textwidth}
			\centering
			\includegraphics[width=\linewidth]{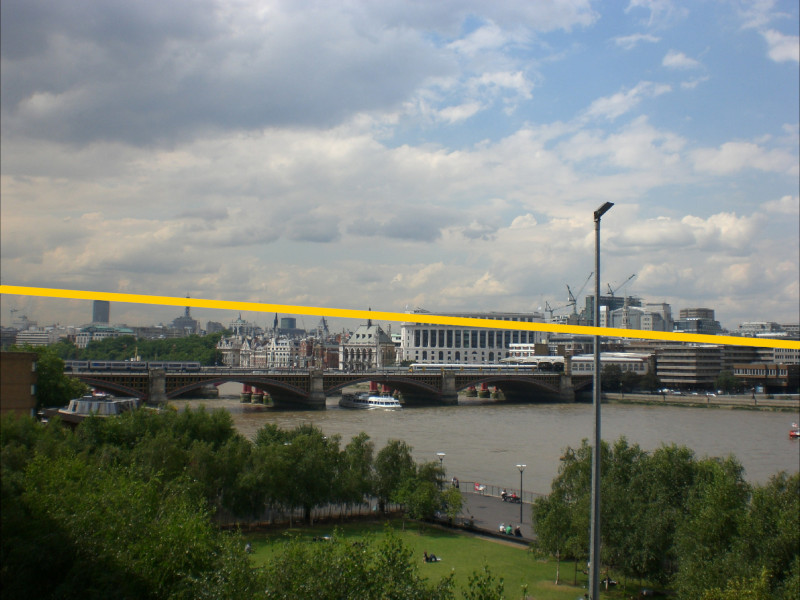}
		\end{subfigure}
		\begin{subfigure}[t]{0.245\textwidth}
			\centering
			\includegraphics[width=\linewidth]{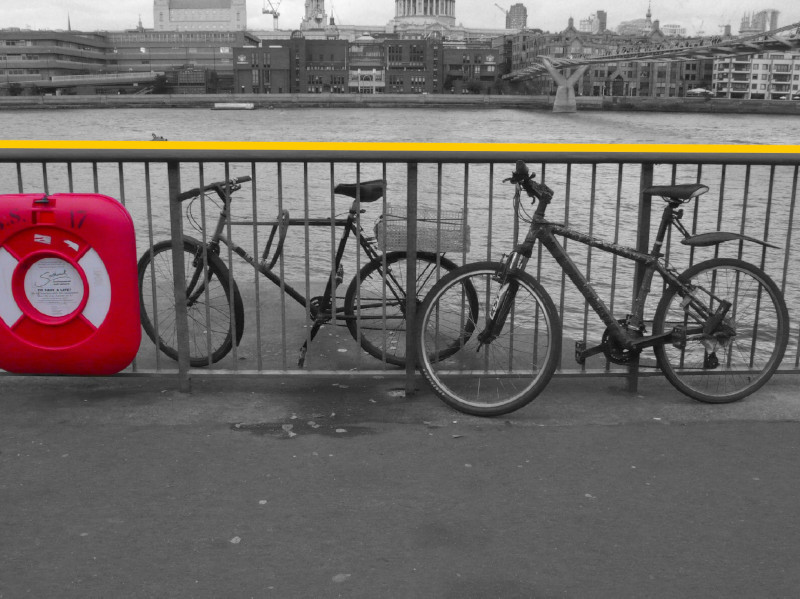}
		\end{subfigure}
		\begin{subfigure}[t]{0.245\textwidth}
			\centering
			\includegraphics[width=\linewidth]{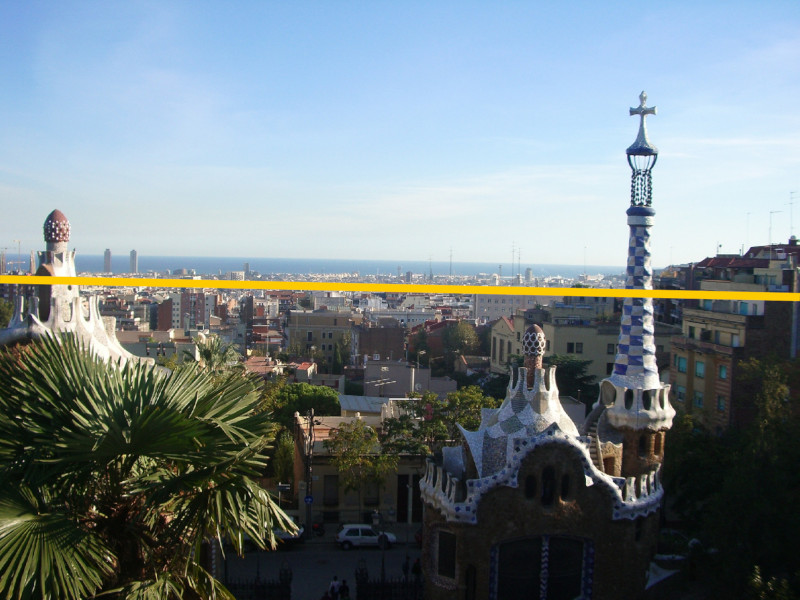}
		\end{subfigure}
		\begin{subfigure}[t]{0.245\textwidth}
			\centering
			\includegraphics[width=\linewidth]{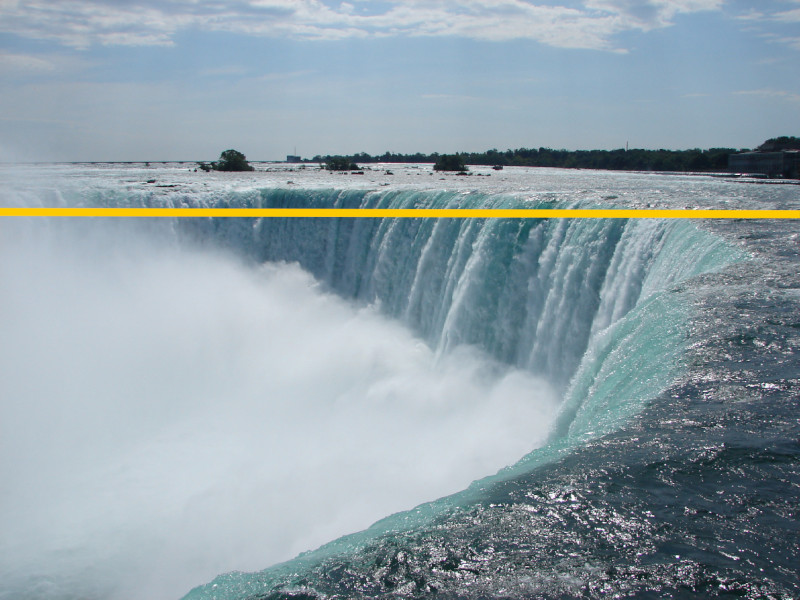}
		\end{subfigure}
		\begin{subfigure}[t]{0.245\textwidth}
			\centering
			\includegraphics[width=\linewidth]{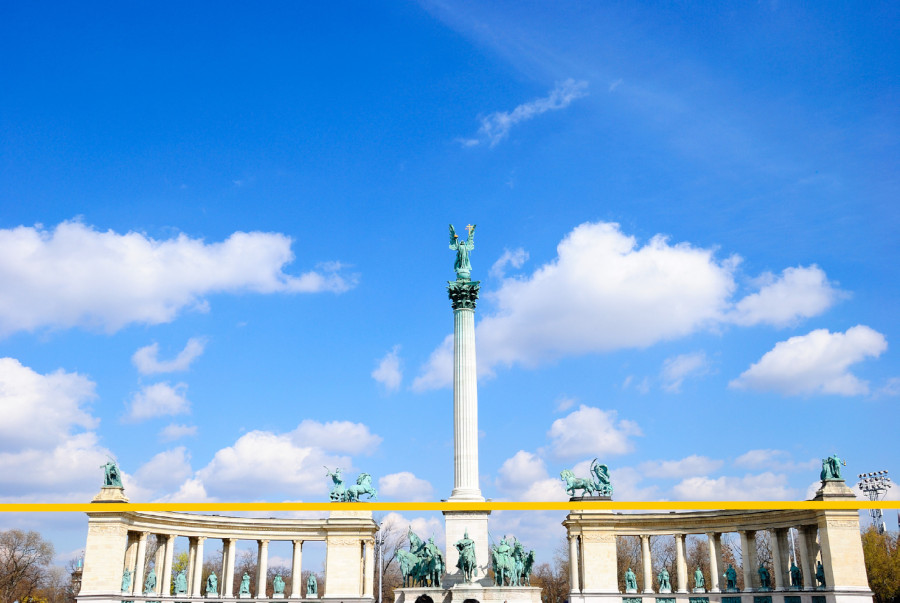}
		\end{subfigure}
		\begin{subfigure}[t]{0.245\textwidth}
			\centering
			\includegraphics[width=\linewidth]{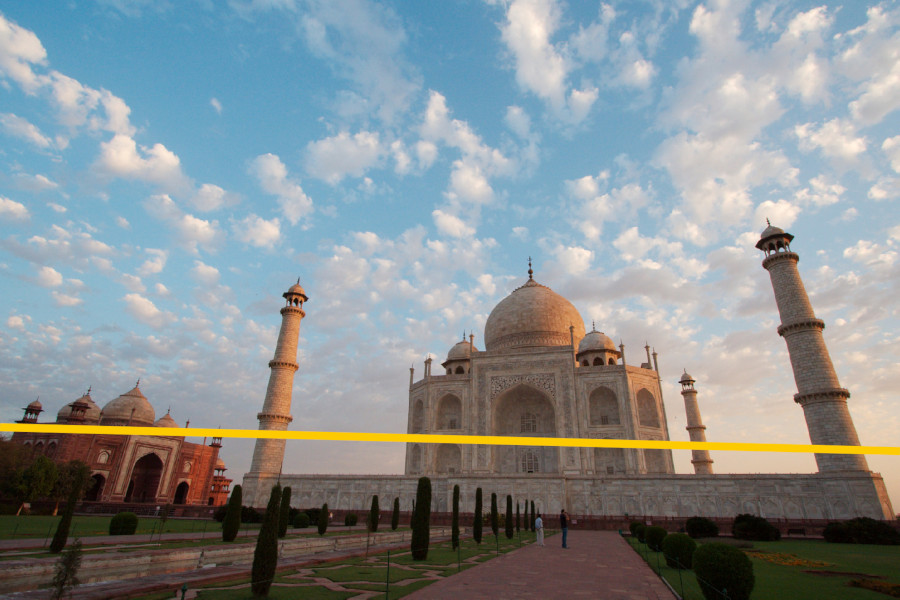}
		\end{subfigure}
		\begin{subfigure}[t]{0.245\textwidth}
			\centering
			\includegraphics[width=\linewidth]{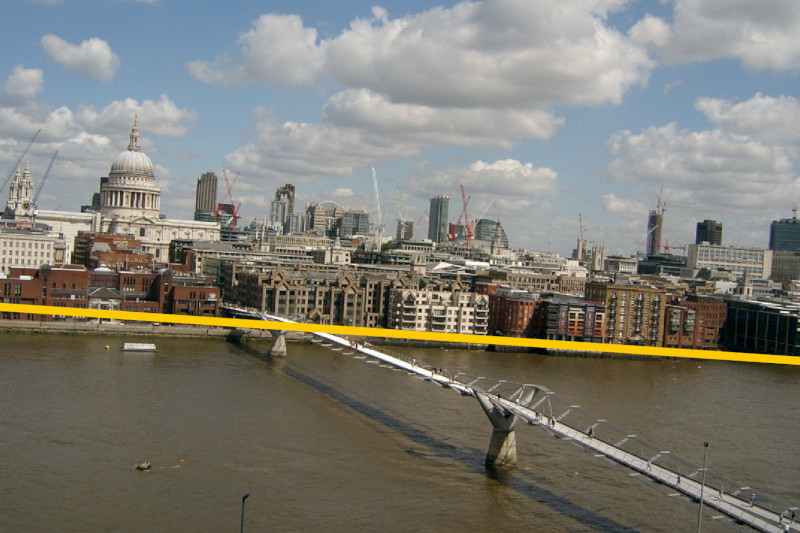}
		\end{subfigure}
		\begin{subfigure}[t]{0.245\textwidth}
			\centering
			\includegraphics[width=\linewidth]{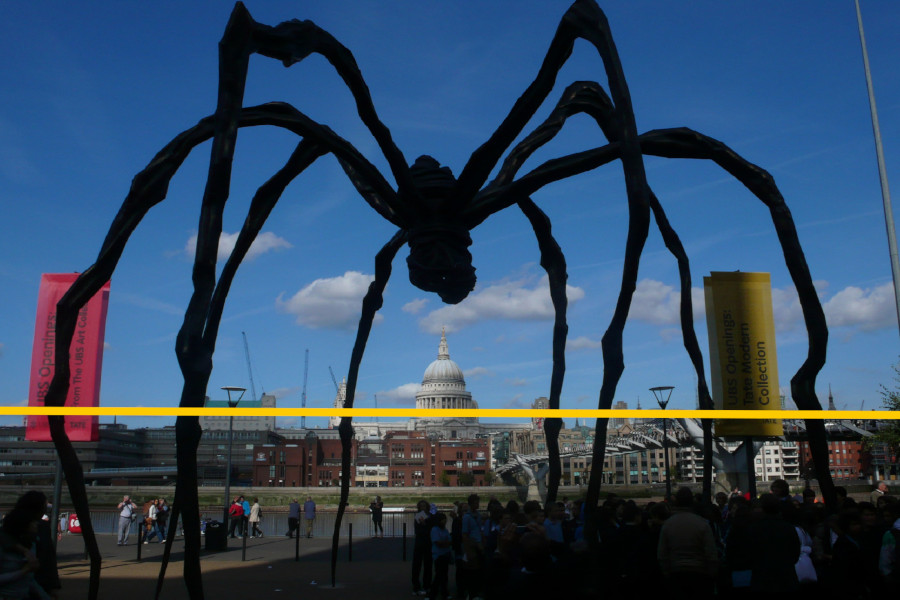}
		\end{subfigure}

		\caption{A selection of images sampled from the HLW dataset. The horizon line labels (\textcolor{niceyellow}{yellow}) are visibly inaccurate.}
		\label{fig:supp_hlw_examples}
	\end{figure*}

As mentioned in the paper, we noticed that some of the horizon line labels provided by the Horizon Lines in the Wild (HLW)~\cite{workman2016hlw} dataset are visibly inaccurate. 
In Fig.~\ref{fig:supp_hlw_examples}, we show a few examples which convey the severity of the problem. 
This is by no means an exhaustive analysis, but we hypothesise that the HLW ground truth contains noticeable errors, which should be kept in mind when using this dataset. 
However, a more detailed analysis is required to quantify and test our hypothesis.

\end{appendices}
\newpage
\clearpage
	{\small
		\bibliographystyle{ieee}
		\bibliography{arxiv}
	}
\end{document}